\documentclass{acmtrans2m}
\usepackage{amsmath}
\usepackage{amssymb}
\usepackage{graphicx}
\usepackage{ifpdf}
\usepackage{color}
\usepackage{algorithm2e}
\usepackage{longtable,lscape}
\usepackage{fancybox}
\newcommand{\nop}[1]{}

\newtheorem{defn}{Definition}

\newtheorem{theorem}{Theorem}
\newtheorem{lemma}{Lemma}
\newtheorem{cor}{Corollary}
\def\qed{\hspace*{\fill} $\blacksquare$}

\title{Theory and Algorithms for Partial Order Based Reduction in Planning}
\author{
  You Xu and Yixin Chen \\ Washington University in St. Louis
  \and 
  Qiang Lu \\ University of Science and Technology of China
  \and
  Ruoyun Huang \\ Washington University in St. Louis
}

\begin{abstract}

Search is a major technique for planning. It amounts 
to exploring a state space of planning domains 
typically modeled as a directed graph.  However, prohibitively 
large sizes of the search space make search expensive. Developing better heuristic functions has been the main technique for improving search efficiency. Nevertheless, recent studies have shown that improving heuristics alone has certain fundamental limits on improving search efficiency. Recently, a new direction of research called partial order based reduction (POR) has been proposed as an alternative to improving heuristics. POR has shown promise in speeding up searches. 

POR has been extensively studied in model checking research and is a key enabling technique for scalability of model checking systems. Although the POR theory has been extensively studied in model checking, it has never been developed systematically for planning before. In addition, the conditions for POR in the model checking theory are abstract and not directly applicable in planning. Previous works on POR algorithms for planning did not establish the connection between these algorithms and existing theory in model checking.

In this paper, we develop a theory for POR in planning. The new theory we develop connects the stubborn set theory in model checking and POR methods in planning. We show that previous POR algorithms in planning can be explained by the new theory. Based on the new theory, we propose a new, stronger POR algorithm. Experimental results on various planning domains show further search cost reduction using the new algorithm.

\end{abstract}

\begin{document}
\maketitle
\section{Introduction}
\label{sec:intro}

State space search is a fundamental and pervasive approach to artificial intelligence in general and planning in particular.  It is among the most successful approaches to planning.  A major concern with state space search is that it has a high time and space cost since the state space that needs to be explored is usually very large. 

Much research on classical planning has focused on the design of better heuristic functions.  For example, new heuristic functions have recently been developed by analyzing the domain transition graphs (DTGs) and causal graphs on top of the SAS+ formalism~\cite{Briel07anlp-based,Helmert08}.  Despite the success of using domain-independent heuristics for classic planning, heuristic planners still face scalability challenges for large-scale problems.  As shown by recent work, search even with almost perfect heuristic guidance may still lead to very high search cost~\cite{Helmert08}.  Therefore, it is important to improve other components of the search algorithm that are orthogonal to the development of heuristics.  

Recently, partial order based reduction (POR), a new way to reduce the search cost from an orthogonal perspective, has been studied for classical planning~\cite{IJCAI09b,IJCAI09a}.  POR as a method to reduce search space has been extensively studied in model checking with solid theoretical investigation.  However, the theoretical properties of POR in planning have still not been fully investigated. There are three key questions. 

1) POR algorithms have been extensively studied in model checking. In fact, POR is an {\em enabling} technique for modeling checking, which will not be practical without POR due to its high time complexity. Extensive research has been developed for the theory of POR in model checking. What are the relationships between the previous POR methods designed for model checking and existing work for planning? Understanding these relationships can not only help us understand both problems better, but can also potentially lead to better POR algorithms for planning. 

2) In essence, all POR based algorithms reduce the search space by restricting certain actions from expanding at each state. Although these POR algorithms all look similar, what are the differences in the quality of reduction that significantly affect search efficiency? We think it is important to investigate the reduction powers of different POR algorithms. 

3) Given the fact that there is more than one POR reduction algorithm for planning, are there other, stronger POR algorithms? To answer this question, in essence, we need to find the sufficient and/or necessary conditions for partial-order based pruning.  There are sufficient conditions for POR in model checking.  Nevertheless, those conditions are abstract and not directly applicable in planning. 

The main contribution of this work is to establish the relationship between the POR methods for model checking and those for planning.  We leverage on the existing POR theory for model checking and develop a counterpart theory for planning.  This new theory allows existing POR algorithms for planning to be explained in a unified framework.  Moreover, based on the conditions given by this theory,  we develop a new POR algorithm for planning that is stronger than previous ones. Experimental results also show that our proposed algorithm leads to more reduction. % and less overhead. 
  
This paper is organized as follows. We first give basic definitions in Section~\ref{sec:background}.  In Section 3, we present a general theory that gives sufficient conditions for POR in planning.  In Section 4, we use the new theory to explain two previous POR algorithms.  Based on the theory, in Section 5, we propose a new POR algorithm for planning which is different and stronger than previous ones.  We report experimental results in Section~\ref{sec:results}, review some related work in Section~\ref{sec:related}, and give conclusions in Section~\ref{sec:conclusion}.

\section{Background}
\label{sec:background}

Planning is a core area of artificial intelligence. It entails arranging a course of actions to achieve certain goals under given constraints. Classical planning is the most fundamental form of planning, which deals with only propositional logic. % logic is not countable
In this paper, we work on the SAS+ formalism~\cite{jonsson&Bm98}
of classical planning. 
SAS+ formalism has recently attracted a lot of attention due to 
a number of advantages it has over the traditional STRIPS formalism.  
In the following, we review this formalism and introduce our notations.

\begin{defn}
A \textbf{SAS+ planning task} $\Pi$ is defined as a tuple of four elements, $\Pi = \{X, \mathcal{O}, S, s_I, s_G\}$.
\begin{itemize}
\item $X = \{ x_{1}, \cdots, x_{N} \}$ is a set of multi-valued {\bf state variables}, each with an associated finite domain $Dom(x_{i})$.

\item $\mathcal{O}$ is a set of actions and each action $o \in \mathcal{O}$ is a tuple $(pre(o), \textit{eff}(o))$, where both $pre(o)$ and $\textit{eff}(o)$ define some partial assignments of state variables in the form $x_{i} = v_{i}, v_{i} \in Dom(x_{i})$. $s_G$ is a partial assignment that defines the goal.

\item $S$ is the set of states. A {\bf state} $s \in S$ is  a full assignment to all the state variables. $s_I \in S$ is the {\bf initial state}. A state $s$ is a {\bf goal state} if $s_G \subseteq s$.
\end{itemize}
\end{defn}

\begin{defn}\label{def.conflict}
Two partial assignment sets are {\bf conflict-free} if and only if they do not assign different values to the same state variable.
\end{defn} 

For a SAS+ planning task, for a given state $s$ and an action $o$, when all variable assignments in $pre(o)$ are  met in state $s$, action $o$ is {\it applicable} in state $s$. After applying $o$ to $s$, the state variable assignment will be changed to a new state $s'$ according to $\textit{eff}(o)$: the state variables that appear in $\textit{eff}(o)$ will be changed to the assignments in $\textit{eff}(o)$ while other state variables remain the same.
We denote the resulting state after applying an applicable action $o$ to $s$ as $s'=apply(s, o)$. $apply(s, o)$ is undefined if $o$ is not applicable
in $s$. The planning task is to find a \textbf{path}, or a
sequence of actions, that transits the initial state $s_I$ to
a goal state that includes $s_G$.

An important structure for a given SAS+ task is the domain transition graph defined as follows:

% XXX: we create a virtual node v0 here to denote add effects without precondition
\begin{defn}
For a SAS+ planning task, each state variable $x_i$ $(i=1, \cdots, N)$  corresponds to a \textbf{domain transition graph (DTG)} $G_i$, a directed graph with a vertex set $V(G_i)=Dom(x_i) \cup v_0$, where $v_0$ is a special vertex, and an edge set $E(G_i)$ determined by the following.
\begin{itemize}
\item If there is an
action $o$ such that $(x_i = v_i) \in pre(o)$ and  $(x_i = v_i') \in \textit{eff}(o)$, then
$(v_i, v_i')$ belongs to $E(G_i)$ and we say that
$o$ is \textbf{associated} with the edge $e_i = (v_i, v_i')$ (denoted as $o
\vdash e_i$). It is conventional to call the edges in DTGs \textbf{transitions}.

\item If there is an
action $o$ such that  $(x_i = v_i') \in \textit{eff}(o)$ and no assignment to $x_i$ is in $pre(o)$, then
$(v_0, v_i')$ belongs to $E(G_i)$ and we say that
$o$ is \textbf{associated} with the transition $e_i = (v_0, v_i')$ (denoted as $o
\vdash e_i$).
\end{itemize}
\end{defn}

Intuitively, a SAS+ task can be decomposed into multiple objects, each corresponding to one DTG, which models the transitions of the possible values of that object.

\begin{defn}
For a SAS+ planning task, an action $o$ is \textbf{associated} with a DTG $G_i$ (denoted as
$o \vdash G_i$) if $\textit{eff}(o)$ contains an assignment to $x_i$.
\end{defn}
% This definition is revised by Eric, previously we define that o is associated with an edge in G, but -1 => f in G should also be considered as associated 

\begin{defn}
For a SAS+ planning task, a DTG $G_i$ is \textbf{goal-related} if the partial
assignments in $s_G$ that define the goal states include an assignment $x_i = g_i$ in
$G_i$. A goal-related DTG is {\bf unachieved} in state $s$ if $x_i = v_i$ in $s$ and
$v_i \neq g_i$. 
\end{defn}

A SAS+ planning task can also specify a \textbf{preference} that needs to be optimized. A preference is a mapping from a path $p$ to a numerical value.
In this paper we assume an \textbf{action set invariant preference}.
A preference is action set invariant if two paths have the same preference whenever they contain the same set of actions (possibly in different orders).
Most popular preferences, such as plan length and total action cost, are action set invariant.

\section{Partial Order Reduction Theory for Planning}
\label{sec:theory}

Partial order based reduction (POR) algorithms have been extensively studied
for model checking~\cite{Varpaaniemi05,Clarke00}, which also requires examining a state space in 
order to prove certain properties.  POR is a technique that allows a search to explore only part 
of the entire search space and still maintain completeness and/or optimality.
Without POR, model checking would be too expensive to be practical~\cite{Holzmann97}. However, 
POR has not been studied systematically for planning.

In this section, we will first introduce the concept of search reduction. Then, we will present a 
general POR theory for planning, which gives sufficient conditions that guide the design of practical POR algorithms.

\subsection{Search reduction for planning}

We first introduce the concept of search reduction. A standard search,
such as breath-first search (BFS), depth-first search, or $A^*$ search, needs
to explore a state space graph. A reduction algorithm is an algorithm that
reduces the state space graph into a subgraph, so that a search
will be performed on the subgraph instead of the original one. We first define the state space graph. In our presentation, for any graph $G$, we use $V(G)$ to denote the set of vertices and $E(G)$ the set of edges. For a directed graph $G$, for any vertex $s \in V(G)$, a vertex $s' \in V(G)$ is its \textbf{successor} if and only if $(s,s') \in E(G)$.

For a SAS+ planning task, a \textbf{state space graph} for the task is a directed graph $\mathcal{G}$ in which each state $s$ is a vertex and each directed edge $(s,s')$ represents an action that will be explored during a search process.
Most search algorithms work on the original state space graph as defined below.

\begin{defn}
For a SAS+ planning task, its \textbf{original state space graph} is a directed 
graph $\mathcal{G}$ in which each state $s$ is a vertex and there is a directed edge $(s,s')$ if and only 
if there exists an action $o$ such that $apply(s,o) = s'$. We say that action $o$ \textbf{marks} 
the edge $(s,s')$.
\end{defn}

\begin{defn}
For a SAS+ planning task, for a state space graph $\mathcal{G}$, the \textbf{successor set} of a 
state $s$ , denoted by
$succ_{\mathcal{G}}(s)$, is the set of all the successor states of $s$.
The \textbf{expansion set} of a state $s$, denoted by
$expand_{\mathcal{G}}(s)$, is the set of actions
 $$expand_{\mathcal{G}}(s) = \{o~|~o~\textrm{marks}~ (s,s'),~(s,s') \in E(\mathcal{G})\}.$$
\end{defn}

Intuitively, the successor set of a state $s$ includes all the successor states that shall be 
generated by a search upon expanding $s$, while the expansion set includes all
the actions to be expanded at $s$.

In general, a \textbf{reduction method} is a method that 
maps each input state space graph $\mathcal{G}$ to a subgraph of 
$\mathcal{G}$. The POR algorithms we study remove edges from 
$\mathcal{G}$. More specifically, each state $s$ is only 
connected to a subset of all
its successors in the reduced subgraph. 
We note that, by removing edges, a POR algorithm may also 
reduce the number of vertices that are reachable from the initial state, 
hence reducing the number of nodes examined by a search.  
The decision whether a successor state $s'$ would still be 
a successor in the reduced subgraph can be made locally by 
checking certain conditions related to the current state and 
some precomputed information. Hence, a POR algorithm can be combined with 
various search algorithms.

For a SAS+ task, a \textbf{solution sequence} in its state space 
graph $\mathcal{G}$ is a pair $(s^0, p)$, where $s^0$ 
is a non-goal state, $p= (a_1, \dots, a_k)$ is a 
sequence of actions, and, let $s^i = apply(s^{i-1},a_i), i = 1, \dots, k$, $(s^{i-1}, s^i)$ is 
an edge in $\mathcal{G}$ for $i = 1, \dots, k$ and 
$s^k$ is a goal state. We now define some generic properties of reduction methods.

\begin{defn}
For a SAS+ planning task, a reduction method is \textbf{completeness-preserving} if 
for any solution sequence $(s^0, p)$ in the state space graph, 
there also exists a solution sequence $(s^0, p')$ in 
the reduced state space graph. 
\end{defn}

\begin{defn}
For a SAS+ planning task, a reduction method is \textbf{optimality-preserving} if, for any 
solution sequence $(s^0, p)$ 
in the state space graph, there also exists a solution sequence $(s^0, p')$ in the 
reduced state space graph satisfying that $p'$ has the same preference that $p$ does.
\end{defn}

\begin{defn}
For a SAS+ planning task, a reduction method is \textbf{action-pre-serving} if, for any solution 
sequence $(s^0, p)$ 
in the state space graph, there also exists a solution sequence $(s^0, p')$ in 
the reduced state space graph satisfying that the actions in $p'$ is a permutation of the actions 
in $p$.
\end{defn}

Clearly, being action-preserving is a sufficient condition for being completeness-preserving. 
When the preference is action set invariant, being action-preserving is also a sufficient condition
for being optimality-preserving.

\subsection{Stubborn set theory for planning}
Although there are many variations of POR methods, a popular and representative POR algorithm is the stubborn set method~\cite{Valmari88,Valmari91a,Valmari90,Valmari98,Valmari91b,Valmari93}, used for model checking based on Petri nets. The basic idea is to form a stubborn set of applicable actions for each state and expand only the actions in the stubborn set during search.  By expanding a small subset of applicable actions in each state, stubborn set methods can reduce the search space without compromising completeness.

Since planning also examines a large search space, we propose to develop a  stubborn set theory for planning. To achieve this, we need to handle various subtle issues arising from the differences between model checking and planning. We first define the concept of stubborn sets for planning, 
adapted from the concepts in model checking.

\begin{defn}[\textbf{Stubborn Set for Planning}]\label{def.stubborn}
For a SAS+ planning task, a set of actions $T(s)$ is a stubborn set at state $s$  if
and only if
\begin{itemize}
\item[A1)] For any action $b \in T(s)$ and actions $b_1, \cdots,
b_k \notin T(s)$, if $(b_1, \cdots, b_k, b)$ is a prefix of a path from $s$ to a goal state, 
then $(b, b_1, \cdots, b_k)$
is a valid path from $s$ and leads to the same state that $(b_1, \cdots, b_k, b)$ does; and

\item[A2)] Any valid path from $s$ to a goal state contains at least one action in $T(s)$.
\end{itemize}
\end{defn}

\begin{figure}[t]
  \centering
  \scalebox{0.6}{\includegraphics{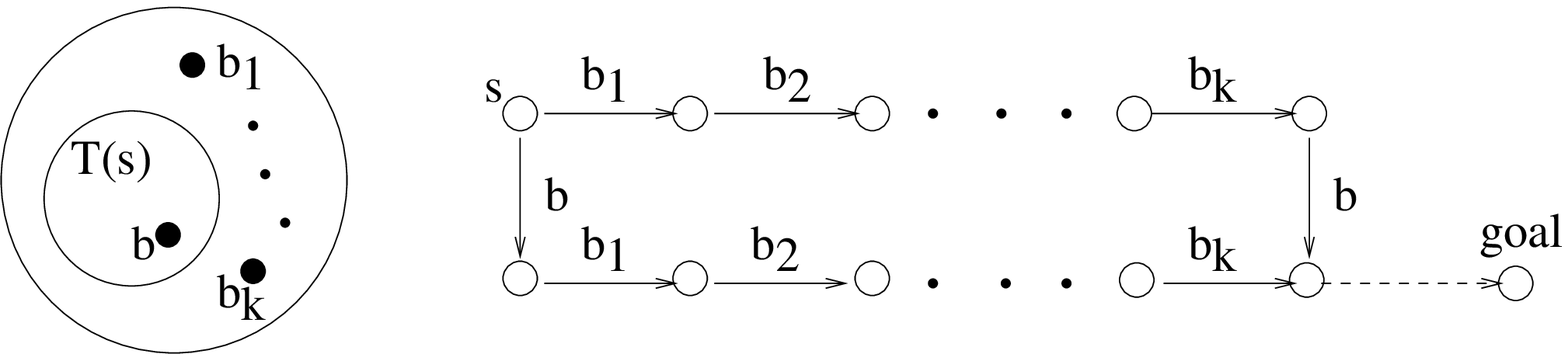}}\\
    This diagram plots the stubborn set condition A1 in Definition~\ref{def.stubborn}. 
    \caption{\label{fig.concept} Illustration of stubborn set.}
\end{figure}

The above definition is schematically illustrated in Figure~\ref{fig.concept}.
Once we define the stubborn set $T(s)$ at each state $s$, we in effect reduce
the state space graph to a subgraph: only the edges corresponding to actions
in the stubborn sets are kept in the subgraph.

\begin{defn}\label{def.expandr.ss}
For a SAS+ planning task, given a stubborn set $T(s)$ defined at each state $s$,
the stubborn set method reduces its state space
graph $\mathcal{G}$ to a subgraph $\mathcal{G}_r$ such that
$V(\mathcal{G}_r) = V(\mathcal{G})$ and there is an edge $(s,s')$ in
$E(\mathcal{G}_r)$ if and only if there exists an action $o \in T(s)$
such that $s' = apply(s,o)$.
\end{defn}

A \textbf{stubborn set method for planning} is a reduction method that
reduces the original state space
graph $\mathcal{G}$ to a subgraph $\mathcal{G}_r$ according to Definition~\ref{def.expandr.ss}.
In other words, a stubborn set method expands actions only in a stubborn set in each state. In 
the sequel, we show that such a reduction method preserves actions,
hence, it also preserves completeness and optimality. 

\begin{theorem}~\label{theo.ap}
Any stubborn set method for planning is action-preserving.
\end{theorem}

\proof
We prove that for any solution sequence
$(s^0,p)$ in the original state space graph $\mathcal{G}$, there exists a
solution sequence $(s^0,p')$ in the reduced state space graph
$\mathcal{G}_r$ resulting from the stubborn set method,
such that $p'$ is a permutation of actions in $p$.
We prove this fact by induction on $k$, the length of $p$. 

When $k=1$, let $a$ be the only action in $p$, according to the second condition in
Definition~\ref{def.expandr.ss}, $a$ is in $T(s_0)$.
Thus, $(s^0, p)$ is also a
solution sequence in $\mathcal{G}_r$. The EC method is
action-preserving in the base case.

When $k>1$, the induction assumption is that any path in $\mathcal{G}$ with length less than 
or equal to $k-1$ has a permutation in $\mathcal{G}_r$ that leads to the same final state. % TODO fix the proof here
Now we consider a solution sequence $(s^0, p)$ in
$\mathcal{G}$: $p=(a_1, \dots, a_k)$. Let $s^i = apply(s^{i-1},a_i), i = 1, \dots, k$.
If $a_1 \in T(s)$, we can invoke the induction assumption for the
state $s^1$ and prove our induction assumption for $k$. % TODO

We now consider the case where $a_1 \notin T(s)$.
Let $a_j$ be the first action in $p$ such that $a_j \in T(s)$. Such an
action must exist because of the condition A2 in
Definition~\ref{def.stubborn}.

Consider the sequence $p^{*} = (a_j,
a_1, \cdots, a_{j-1}, a_{j+1}, \cdots, a_k)$.
According to condition A1 in Definition~\ref{def.expandr.ss}, $(a_j,
a_1, \cdots, a_{j-1})$ is also a valid sequence from $s_0$ which leads to
the same state that $(a_1, \cdots, a_{j})$ does. Hence, we know that $(s^0,p^*)$ is also a 
solution path. Therefore, let $s' = apply(s^0, a_j)$, we know $(a_1, \cdots, a_{j-1})$ is an 
executable action sequence starting from $s'$.
Let $p^{**} = (a_1, \cdots, a_{j-1}, a_{j+1},
\cdots, a_k)$, $(s',p^{**})$ is a solution sequence in
$\mathcal{G}$. From the induction assumption, we know there is a
sequence $p'$ which is a permutation of $p^{**}$, such that
$(s',p')$ is a solution sequence in $\mathcal{G}_r$. Since $a_j \in
T(s^0)$, we know that $a_j$ followed by $p'$ is a solution
sequence from $s^0$ and is a permutation of actions in $p^*$, which is a
permutation of actions in $p$. Thus, the stubborn set method
is action-preserving.
\endproof
% \hspace*{\fill} $\blacksquare$

Since being action-preserving is a sufficient condition for being completeness-preserving 
and optimality-preserving, when the preference is action set invariant, we have the following 
result.

\begin{cor}~\label{cor.p1}
A stubborn set method for planning is completeness-preserving. In addition, it is 
optimality-preserving when the preference is action set invariant.
\end{cor}

\subsection{Left commutativity in SAS+ planning}
% never use   A whaever noun is whatever ... pattern
Note that although Theorem 1 provides an important result for reduction, it is 
not directly applicable since the conditions in Definition~\ref{def.stubborn} are abstract 
and not directly implementable in algorithms.  We need to find sufficient conditions for 
Definition~\ref{def.stubborn} that can facilitate the design of reduction algorithms. In the 
following, we define several concepts that can lead to sufficient conditions for 
Definition~\ref{def.stubborn}.

\begin{defn}[\textbf{State-Dependent Left Commutativity}]\label{def.sdlc}
For a SAS+ planning task, an ordered action pair $(a,b), a,b \in O$ is
left commutative in state $s$, if $(a, b)$ is
a valid path at $s$, and $(b, a)$ is also a valid path at $s$ and
results in the same state. We denote such a relationship by $s: b \Rightarrow a$.
\end{defn}

\begin{defn}[\textbf{State-Independent Left Commutativity}]\label{def.silc}
For a SAS+ 
planning task, an ordered action pair $(a,b), a,b \in O$ is
left commutative if, for any state $s$, it is true that $s: b \Rightarrow a$.
We denote such a relationship by $b \Rightarrow a$.
\end{defn}

Note the following. 1) Left commutativity is not a symmetric
relationship. $b \Rightarrow a$ does not imply $a \Rightarrow b$.
2) The order in the notation $b \Rightarrow a$ suggests that we should always
try only $(b,a)$ during the search instead of trying both $(a,b)$
and $(b,a)$. Also, not every state-independent left commutative action pair 
is state-dependent left commutative. For instance, in a SAS+ planning task
with three state variables $\{x_1, x_2, x_3\}$, action $a$ with $pre(a) = \{ x_1=0 \}$, 
\textit{eff}$(a) = \{ x_2 = 1 \}$ and action $b$ with $pre(b) = \{x_2 = 1, x_3 =2 \}$, 
\textit{eff}$(b) = \{x_3 = 3\}$ are left commutative in state $s_1 = \{x_1 = 0, x_2 = 1, x_3 = 3\}$
but not in state $s_2 = \{x_1 = 0, x_2 = 0, x_3 =2 \}$ as $b$ is not applicable in state $s_2$.

We introduce state-independent left commutativity as it can be used to derive sufficient 
conditions for finding stubborn sets.

\begin{defn}[\textbf{State-Independent Left Commutative Set}]\label{def.left}
For a SAS+ \\ 
planning task, a set of actions $T(s)$ is a left commutative set at a state $s$  if and 
only if

\begin{itemize}
\item[L1)] For any action $b \in T(s)$ and any action $a \in O - T(s)$, if there exists a valid 
path from $s$ to a goal state that contains both $a$ and $b$, then it is the case 
that $b \Rightarrow a$; and

\item[A2)] Any valid path from $s$ to a goal state contains at least one action in $T(s)$.
\end{itemize}
\end{defn}

\begin{figure}[t]
\centering
  \scalebox{0.6}{\includegraphics{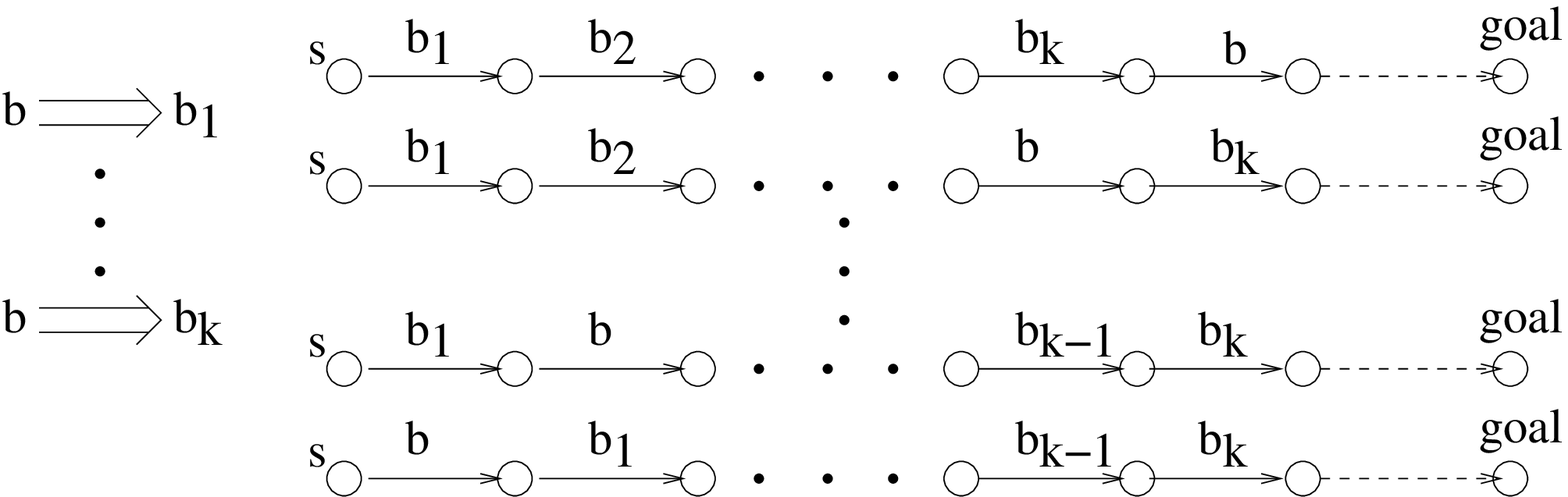}}\\
  \vspace*{0.05in}
    In this diagram, the left part plots the condition L1 in Definition~\ref{def.left} 
         and the right part plots the strategy in the proof to Theorem~\ref{theo.left}. 
   \caption{\label{fig.concept2} Illustration of left commutative set.}
\end{figure}

\begin{theorem}\label{theo.left}
For a SAS+ planning task, for a state $s$, if a set of actions $T(s)$ is a state-independent 
left commutative set, it is also a stubborn set.
\end{theorem}

\proof
We only need to prove that L1 in Definition~\ref{def.left} implies A1 in Definition
~\ref{def.stubborn}. The proof strategy is schematically shown in Figure~\ref{fig.concept2}.

For an action $b \in T(s)$ and actions $b_1, \cdots,
b_k \notin T(s)$, if $(b_1, \cdots, b_k, b)$ is a prefix of a path from $s$ to a goal state, then 
according to L1, we see that $b \Rightarrow b_i$, for $i=1, \cdots, k$. According to the 
definition of left commutativity, we see that $b_k$ and $b$ can be swapped and that the resulting 
path  $(b_1, \cdots, b, b_k)$ is still a valid path that leads to the same state that $(b_1, \cdots, b_k, b)$ does. We can subsequently swap $b$ with $b_{k-1}$, $\cdots$, and $b_1$ to 
obtain equivalent paths, before finally obtaining $(b, b_1, \cdots, b_k)$, as shown in the schematic illustration in the right part of Figure~\ref{fig.concept2}. Hence, we have shown that if 
$p=(b_1, \cdots, b_k, b)$ is a prefix of a path from $s$ to a goal state, then 
$p'=(b, b_1, \cdots, b_k)$ is a also valid path from $s$ that leads to the same state that $p$ does, 
which is exactly the condition $A1$ in Definition~\ref{def.stubborn}.
\endproof

From the above proof, we see that the requirement of state-independent left commutativity in 
Definition~\ref{def.left} is unnecessarily strong. Instead, only certain state-dependent left 
commutativity is necessary. In fact, when we change 
$(b_1, \cdots, b_k, b)$ to $(b_1, \cdots, b, b_k)$, we only require $s': b \Rightarrow b_k$
where $s'$ is the state after $b_{k-1}$ is executed. Similarly, when we change 
$(b_1, \cdots, b_k, b)$ to  $(b_1, \cdots, b, b_{k-1}, b_k)$, we only require 
$s'': b \Rightarrow b_{k-1}$ where $s''$ is the state after $b_{k-2}$ is executed. Based on the 
above analysis, we can refine the sufficient conditions. 
% obtain the following results with weaker requirements.

\begin{defn}[\textbf{State-Dependent Left Commutative Set}]\label{def.left2}
For a SAS+ \\ 
planning task, a set of actions $T(s)$ is a left commutative set at a state $s$  if and
only if
\begin{itemize}

    \item[L1')] For any action $b \in T(s)$ and actions $b_1, \cdots, b_k \notin T(s)$, 
if $(b_1, \cdots, b_k, b)$ is a prefix of a path from $s$ to a goal state, then 
$s':b \Rightarrow b_k$, where $s'$ is the state after $(b_1, \cdots, b_{k-1})$ is executed; and

    \item[A2)] Any valid path from $s$ to a goal state contains at least one action in $T(s)$.
\end{itemize}

\end{defn}

We only need to slightly modify the proof to Theorem~\ref{theo.left} in order to prove the 
following theorem.

\begin{theorem}\label{theo.left2}
For a SAS+ planning task, for a state $s$, if a set of actions $T(s)$ is a state-dependent left 
commutative set, it is also a stubborn set.
\end{theorem}

The above result gives sufficient conditions for finding stubborn sets in planning. The 
concept of state-dependent left commutative set requires a less stringent condition than the 
state-independent left commutative set. 
Such a nuance actually leads to different previous POR algorithms with varying performances. 
Therefore, it will result in smaller $T(s)$ sets and 
stronger reduction.  Next, we present our algorithm for finding such a set at each state to satisfy these conditions.

\subsection{Determining left commutativity}

Theorem~\ref{theo.left2} provides a key result for POR. However, the conditions in 
Definition~\ref{def.sdlc} are still abstract and not directly implementable.
The key issue is to efficiently find left commutative action pairs. 
Now we give necessary and sufficient conditions for Definition~\ref{def.sdlc} that can 
practically determine left commutativity and facilitate the design of reduction algorithms. 

\begin{theorem} \label{theo.comm}
For a SAS+ planning task, for a valid action path $(a, b )$ in state $s$, 
we have $s: b \Rightarrow a$ if and only if $pre(a)$ and $\textit{eff}(b)$, $pre(b)$ and $\textit{eff}(a)$,
$\textit{eff}(a)$ and $\textit{eff}(b)$ are all conflict-free and $b$ is applicable
at $s$.
\end{theorem}

\proof 
First, from the definition of $s: b \Rightarrow a$, we know that 
action $b$ is applicable in state $apply(s, a)$. This implies that $pre(b)$ and \textit{eff}$(a)$ 
are conflict-free.  Symmetrically, since action $a$ is applicable in state 
$apply(s, b)$, $pre(a)$ and \textit{eff}$(b)$ are also conflict-free.
Now we prove \textit{eff}$(a)$ and \textit{eff}$(b)$ are conflict-free by contradiction.  
If \textit{eff}$(a)$ and \textit{eff}$(b)$ are not conflict-free, without loss of generality, we can % you need a 'can' here
assume that $\textit{eff}(a)$ contains $x_i = v_i$ and $\textit{eff}(b)$ 
contains $x_i = v_i' \neq v_i$.  Thus, the value of $x_i$ is $v_i$ for state 
$s_{ab} = apply(apply(s, a), b)$ and $v_i'$ for state $s_{ba} = apply(apply(s, b), a)$, 
i.e., $s_{ab}$ is different than $s_{ba}$. This contradicts our assumption that $a$ and $b$ are 
left commutative. Thus, $\textit{eff}(a)$ and $\textit{eff}(b)$ are conflict-free. 

Second, if $pre(a)$ and $\textit{eff}(b)$, $\textit{eff}(a)$ 
and $pre(b)$, $\textit{eff}(a)$ and $\textit{eff}(b)$ are all conflict-free, 
since $a$ is applicable in $s$, $a$ is also applicable in state $apply(s, b)$ 
as $pre(a)$ and \textit{eff}$(b)$ are conflict-free. Hence,  $(b, a)$ is a valid path at $s$.
Also, for any state variable $x_i$, its value in states $s_{ab} = apply(apply(s, a), b)$ and 
$s_{ba} = apply(apply(s, b), a)$ are the same, because
\textit{eff}$(a)$ and \textit{eff}$(b)$ are conflict-free.  Therefore, we have $s_{ab} = s_{ba}$. 
Hence, we have $s: b \Rightarrow a$. 
\endproof

Theorem~\ref{theo.comm} gives necessary and sufficient
conditions for deciding whether two actions are left-commutative or not. Based 
on this result, we later develop practical POR algorithms that find stubborn 
sets using left commutativity.

\section{Explanation of previous POR algorithms}
\label{sec:prepor}

Previously, we have proposed two POR algorithms for planning: expansion core
(EC)~\cite{IJCAI09a} and stratified planning (SP)~\cite{IJCAI09b}, both of 
which showed good performance in reducing the search space. However we did 
not have a unified explanation for them.  We now explain how these two 
algorithms can be explained by our theory.
Full details of the two algorithms can be found in our
papers~\cite{IJCAI09a,IJCAI09b}.

\subsection{Explanation of EC}

Expansion core (EC) algorithm is a POR-based reduction algorithm for planning. We will see that, in essence, the EC algorithm exploits the SAS+ formalism to find a left commutative set for each state. To describe the EC algorithm, we need the following definitions. 

\begin{defn}
For a SAS+ task, for each DTG $G_i, i=1, \dots, N$, for a vertex
$v \in V(G_i)$, an edge $e \in E(G_i)$ is a \textbf{potential
descendant edge} of $v$ (denoted as $v \lhd e$) if 1) $G_i$ is
goal-related and there exists a path from $v$ to the goal state in
$G_i$ that contains $e$; or 2)  $G_i$ is not goal-related and $e$
is reachable from $v$.
\end{defn}

\begin{defn}
For a SAS+ task, for each DTG $G_i, i=1, \dots, N$, for a vertex
$v \in V(G_i)$, a vertex $w \in V(G_i)$ is a
\textbf{potential descendant vertex} of $v$ (denoted as $v \lhd
w$) if 1) $G_i$ is goal-related and there exists a path from $v$
to the goal state in $G_i$ that contains $w$; or 2)  $G_i$ is not
goal-related and $w$ is reachable from $v$.
\end{defn}

\begin{figure*}[t]
  \centering
  \scalebox{0.7}{\includegraphics{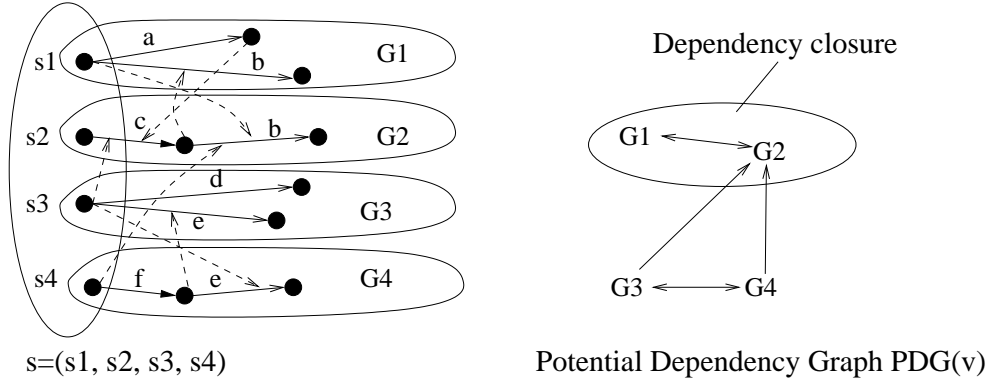}}
   \caption{\label{fig.expand}\small A SAS+ task with four DTGs.
   The dashed arrows show preconditions (prevailing and transitional) of each edge (action). Actions are marked with letters a to f. We see that b and e are associated with more than one DTG.}
   \vspace*{-0.1in}
\end{figure*}

\begin{defn}
For a SAS+ task, given a state $s=(s_1, \cdots, s_N)$, for any $1 \le i,j \le N, i
\neq j$, we call $s_i$ a \textbf{potential precondition} of the DTG $G_j$
if there exist $o \in \mathcal{O}$ and $e_j \in E(G_j)$ such that
\begin{eqnarray}
s_j \lhd e_j, ~o \vdash  e_j,~\textrm{~and~} s_i \in pre(o)
\end{eqnarray}
\end{defn}

\begin{defn}
For a SAS+ task, given a state $s = (s_1, \dots, s_N)$, for any $1 \le i,j \le N, i
\neq j$, we call $s_i$ a \textbf{potential dependent} of the DTG $G_j$ if
there exists $o \in \mathcal{O}$, $e_i=(s_i,s'_i) \in E(G_i)$ and
$w_j \in V(G_j)$ such that
\begin{eqnarray}
s_j \lhd w_j,  ~o \vdash  e_i,~\textrm{~and~} w_j \in pre(o)
\end{eqnarray}
\end{defn}

\begin{defn}~\label{def.pdg}
For a SAS+ task, for a state $s = (s_1, \dots, s_N)$,
its \textbf{potential dependency graph} PDG($s$) is a
directed graph in which each DTG $G_i, i=1, \cdots,
N$ corresponds to a vertex, and there is an edge from $G_i$ to
$G_j$, $i \neq j$, if and only if $s_i$ is a potential
precondition or potential dependent of $G_j$.
\end{defn}

Figure~\ref{fig.expand} illustrates the above definitions. In
PDG($s$), $G_1$ points to $G_2$ as $s_1$ is a potential
precondition of $G_2$ and $G_2$ points to $G_1$ as $s_2$ is a
potential dependent of $G_1$.

\begin{defn}~\label{def.dc}
For a directed graph $H$,  a subset $C$ of $V(H)$ is a
\textbf{dependency closure} if there do not exist $v \in C$ and $w
\in V(H) - C$ such that $(v,w) \in E(H)$.
\end{defn}

Intuitively, a DTG in a dependency closure may depend on other DTGs in
the closure but not those DTGs outside of the closure.
In Figure~\ref{fig.expand}, $G_1$ and $G_2$
form a dependency closure of PDG($s$).

The EC algorithm is defined as follows:

\begin{defn}[\textbf{Expansion Core Algorithm}]\label{def.expandr}
For a SAS+ planning task, the EC method reduces its state space graph $\mathcal{G}$ to a
subgraph $\mathcal{G}_r$ such that $V(\mathcal{G}_r) = V(\mathcal{G})$ and for each
vertex (state) $s \in V(\mathcal{G})$, it expands actions in the following set $T(s) \subseteq O$:
\begin{eqnarray}\label{eqn.expand2}
T(s) = \bigcup_{i \in {\mathcal C}(s)} \bigg\{
o \bigg| o \in exec(s) \wedge o \vdash G_i \bigg\},
\end{eqnarray}
where $exec(s)$ is the set of executable actions in $s$ and
${\mathcal C}(s) \subseteq \{1, \cdots, N\}$ is an index set
satisfying:
\begin{itemize}
\item[EC1)] The DTGs $\{G_i, i \in {\mathcal C}(s)\}$ form a
dependency closure in PDG($s$); and
\item[EC2)] There exists $i \in {\mathcal C}(s)$ such that $G_i$ is goal-related and $s_i$ is not the goal state in $G_i$.

%\item[3)] there exists at least one subgraph $G_i, i \in {\mathcal %C}(v)$, such that $expand(v_i) \neq \varnothing$.
\end{itemize}
\end{defn}

Intuitively, the EC method can be described as
follows. To reduce the original state-space graph, for each state,
instead of expanding actions in all the DTGs, it only expands actions in DTGs that belong to a dependency closure of PDG($s$) under the condition
that at least one DTG in the dependency closure is goal-related and not
at a goal state.

The set ${\mathcal C}(s)$ can always be found for any non-goal
state $s$ since PDG($s$) itself is always such a dependency
closure. If there is more than one such closure,
theoretically any dependency closure satisfying the above
conditions can be used in EC.
In practice, when there are multiple such
dependency closures, EC picks the one with less actions
in order to get stronger reduction. EC has adopted the following scheme 
to find the dependency closure for any state $s$.

Given a PDG($s$), EC first finds its strongly connected components % never say of it
(SCCs). If each SCC is contracted to a single vertex, the
resulting graph is a directed acyclic graph ${\mathcal S}$. Note 
that each vertex in ${\mathcal S}$ with a zero out-degree
corresponds to a dependency closure. It then topologically sorts all the
vertices in ${\mathcal S}$ to get a sequence of SCCs: $S_1, S_2, \cdots$,
and picks the minimum $m$ such that $S_m$ includes a goal-related DTG
that is not in its goal state. It chooses all the DTGs in $S_1, \cdots, S_m$
as the dependency closure.

Now we explain the EC algorithm using the POR theory we developed in Section~\ref{sec:theory}. 
We show that the EC algorithm can be viewed as an algorithm for finding a state-dependent 
left-commutative set in each state. 

\begin{lemma}~\label{theo.p1}
For a SAS+ planning task, the EC algorithm defines a state-dependent left commutative set for each state.
\end{lemma}

\proof Consider the set of actions $T(s)$ expanded by the EC algorithm in each state $s$, as defined in (\ref{eqn.expand2}). We prove that $T(s)$ satisfies conditions L1' 
and A2 in Definition~\ref{def.left2}.

Consider an action $b \in T(s)$ and actions $b_1, \cdots,
b_k \notin T(s)$ such that $(b_1, \cdots, b_k, b)$ is a prefix of a 
path from $s$ to a goal state, we show that $s':b \Rightarrow  b_k$, where $s'$ is the state after $(b_1, \cdots, b_{k-1})$ is applied to $s$.

Let  ${\mathcal C}(s)$  be the index set of the DTGs that form a dependency closure, as used in in (\ref{eqn.expand2}).
Since $b \in T(s)$, there must exist $m \in {\mathcal C}(s)$ such that $b \vdash G_m$.
Let the state after applying $(b_1, \cdots, b_k)$ to $s$ be $s^*$.
We see that we must have $s^*_m = s_m$ because otherwise there must exist a $b_j, 1 \le j \le m$ that changes the assignment of state variable $x_m$. However, that would imply that $b_k \in T(s)$. Since $b$ is applicable in $s^*$, we see that $s_m = s^*_m \in pre(b)$.

If there exists a state variable $x_i$ such that an assignment to $x_i$ is in both \textit{eff}$(b_k)$ and $pre(b)$, then $G_m$ will point to the DTG $G_i$ as $s_m$ is a potential dependent of $G_i$, forcing $G_i$ to be included in the dependency closure, i.e. $i \in {\mathcal C}(s)$. However, as $b_k \vdash G_i$, it will violate our assumption that $b_k \notin T(s)$.
Hence, none of the precondition assignments of $b$ is added by $b_k$. Therefore, since $b$ is applicable in $apply(s',b_k)$, it is also applicable in $s'$.

On the other hand, if $b_k$ has a precondition assignment in a DTG
that $b$ is associated with, then $G_m$ will point to that
DTG since $s_m$ is a potential precondition of $b_k$, forcing that DTG to be in ${\mathcal C}(s)$, which 
contradicts the assumption that $b_k \notin T(s)$. Hence, $b$ does not alter any
precondition assignment of $b_k$. Therefore, since $b_k$ is applicable in $s'$, it is also applicable in the state $apply(s',b)$.

Finally, if there exists a state variable $x_i$ such that an assignment to $x_i$ is altered by both $b$ and $b_k$, then we know $b \vdash G_i$ and $b_k \vdash G_i$.
In this case, $G_m$ will point to $G_i$ since $s_m$ is a potential precondition of $G_i$,
making $b_k \in T(s)$, which contradicts our assumption.
Hence, \textit{eff}$(b)$  and \textit{eff}$(b_k)$ correspond to assignments to distinct sets of state variables. Therefore, applying $(b_k,b)$ and $(b,b_k)$ to $s'$ will lead to the same state.

From the above, we see that $b$ is applicable in $s'$, $b_k$ is applicable in $apply(s',b)$, and hence $(b,b_k)$ is applicable in $s'$. Further we see that $(b,b_k)$ leads to the same state as $(b_k,b)$ does when applied to $s'$. We conclude that $s':b \Rightarrow b_k$ and $T(s)$ satisfies L1'.

Moreover, for any goal-related DTG $G_i$, if in a state $s$, its assignment $s_i$ is not the goal state in $G_i$, then some actions associated with $G_i$ have to be executed in any solution 
path from $s$. Since $T(s)$ includes all the actions in at least one goal-related DTG $G_i$, any solution path must contain at least one action in $T(s)$. Therefore, 
$T(s)$ also satisfies A2 and it is indeed a state-dependent left commutative set.
\endproof

From Lemma~\ref{theo.p1} and Theorem~\ref{theo.left2}, we obtain the following result, which 
shows that EC fits our framework as a stubborn set method for planning. 

\begin{theorem}~\label{theo.p2}
For any SAS+ planning task, the EC algorithm defines a stubborn set in each state.
\end{theorem}

\subsection{Explanation of SP}

The stratified planning (SP) algorithm exploits commutativity of
actions directly~\cite{IJCAI09b}. To describe the SP algorithm, we need the following definitions first. 

\begin{defn}\label{def.cg}
Given a SAS+ planning task $\Pi$ with state variable set $X$, the
\textbf{causal graph ($CG$)} is a directed graph $CG(\Pi) = (X,
E)$ with $X$ as the vertex set. There is an edge $(x, x') \in E$
if and only if $x \neq x'$ and there exists an action $o$ such
that $ x \in \textit{eff}(o)$ and $x' \in pre(o)$ or \textit{eff}$(o)$. 
\end{defn}

\begin{defn} For a SAS+ task $\Pi$, a \textbf{stratification} 
of the causal graph $CG(\Pi)$ as $(X, E)$ is a partition of the node set
$X$: $X = (X_1, \cdots, X_k)$ in such a way that there exists no
edge $e = (x, y)$ where $x \in X_i, y \in X_j$ and $i > j$.
\end{defn}

\nop{
\begin{figure}[t]
    \begin{center}
    \includegraphics[scale=0.5]{ccg.png} \\
        a) Causal graph (CG)\\
    \includegraphics[scale=0.4]{str.png} \\
        b) A stratification of the CG
    \caption{\label{fig.cg.truck} \small
    The causal graph and a stratification of Truck-02. }
    \end{center}
\end{figure}
}

By stratification, each state variable is assigned a level $L(x)$,
where $L(x) = i$ if $x \in X_i, 1 \le i \le k$. Subsequently, each
action $o$ is assigned a level $L(o)$, $1 \le L(o) \le k$. $L(o)$
is the level of the state variable(s) in \textit{eff}$(o)$. Note that all
state variables in the same \textit{eff}$(o)$ must be in the same level, hence, our $L(o)$ is well-defined. 

\begin{defn}[\textbf{Follow-up Action}]\label{followup.action} 
For a SAS+ task $\Pi$, an action $b$ is a follow-up action of $a$
 (denoted as $a \vartriangleright b$) if   $\textit{eff}(a) \cap pre(b) \neq \emptyset $ or $\textit{eff}(a) \cap \textit{eff}(b) \neq \emptyset$.
\end{defn}

The SP algorithm can be combined with standard search algorithms,
such as breadth-first search, depth-first search, and best-first
search (including $A^*$). During the search, for each state $s$
that is going to be expanded, the SP algorithm examines the action
$a$ that leads to $s$. Then, for each
applicable action $b$ in state $S$, SP makes the following
decisions.

\begin{defn}[\textbf{Stratified Planning Algorithm}]\label{def.stra}
For a SAS+ planning task, in any non-initial state $s$, assuming $a$ is 
the action that leads directly to $s$, and $b$ is an applicable 
action in $s$, then SP does not expand $b$ 
if $L(b) < L(a)$ and $b$ is not a follow-up action of $a$. 
Otherwise, SP expands $b$.  In the initial state $s_0$, SP expands all applicable actions.
\end{defn}

\nop{
If $L(b) < L(a)$ and $b$ is not a follow-up action of $a$,
then do not expand $b$ (we say that $b$ is not \textbf{SP
expandable} after $a$). Otherwise, expand $b$.
}

\nop{
\begin{figure}[t]
	\begin{center}
        \includegraphics[scale=0.5]{stratify.png}
	\caption{Stratified Planning Strategy}
	\label{stratify}
	\end{center}
\end{figure}

Figure~\ref{stratify} illustrates the strategy of 
stratified planning algorithm. 
} % end of nop

% \noindent \textbf{Theoretical analysis of SP}
The following result shows the relationship between the 
SP algorithm and our new POR theory. 

\begin{lemma}
\label{sp.left}
If an action $b$ is not SP-expandable after $a$, and state $s$ is the state
before action $a$, then $s: b \Rightarrow a$.
\end{lemma}

\proof
Since $b$ is not SP-expandable after $a$, following the SP algorithm, we have $L(a) >
L(b)$ and $b$ is not a follow-up action of $a$. 
According to Definition~\ref{followup.action},  we have $\textit{eff}(a) 
\cap pre(b) =\textit{eff}(a) \cap \textit{eff}(b)
= \emptyset$. These imply that 
\textit{eff}$(a)$ and $pre(b)$ are conflict-free, and that \textit{eff}$(a)$ 
and $ \textit{eff}(b)$ are conflict-free.  Also, since $b$ is 
applicable in $apply(s, a)$ and \textit{eff}$(a)$ and $pre(b)$ are conflict-free,
$b$ must be applicable in $s$ (Otherwise \textit{eff}$(a)$ must change the value of at least one variable in $pre(b)$, which means 
\textit{eff}$(a)$ and $pre(b)$ are not conflict-free).

Now we prove that $pre(a)$ and \textit{eff}$(b)$ are conflict-free by showing 
$pre(a) \cap \textit{eff}(b) = \emptyset$. If their intersection is non-empty, 
we assume a state variable $x$ is assigned by both $pre(a)$ and \textit{eff}$(b)$. 
By the definition of stratification, $x$ is in layer $L(b)$. 
However, since $x$ is assigned by $pre(a)$, 
% TODO this is incomplete
% edges from variables in \textit{eff}$(a)$ to $x$ in CG. 
% Thus,
there must be an edge from layer $L(a)$ to layer $L(x) = L(b)$ 
since $L(a) \neq L(b)$. In this case, we know 
that $L(a) < L(b)$ from the definition of stratification.
Nevertheless, this contradicts with the assumption 
that $L(a) > L(b)$. Thus, $pre(a) \cap \textit{eff}(b) = \emptyset$, and
$pre(a)$ and \textit{eff}$(b)$ are conflict-free.

With all three conflict-free pairs, we have $s: b \Rightarrow a$ 
according to Theorem~\ref{theo.left}.
\endproof

Although SP reduces the search space by avoiding the expansion 
of certain actions, it is in fact not a stubborn set based reduction algorithm. We have the following theorem for the SP algorithm.

\begin{defn}
For a SAS+ planning task $S$, a valid path $p_a = (a_1, \cdots, a_n)$
is an \textbf{SP-path} if and only if $p_a$ is a path 
in the search space of the SP algorithm applied to S. 
\end{defn}

\begin{theorem}
\label{thm.sp2}
For a SAS+ planning task $S$, for any initial $s_0$ and any valid 
path $p_a = ( a_1, \cdots, a_n) $ from $s_0$, there 
exists a path $p_b = (b_1, \cdots, b_n)$ from $s_0$ such 
that $p_b$ is an SP-path, and both $p_a$ and $p_b$ lead to the same state
from $s_0$, and $p_b$ is a permutation of actions in $p_a$. 
\end{theorem}

\proof
We prove by induction on the number of actions. 

When $n=1$, since there is no action before $s_0$, 
any valid path $(a_1)$ will also be a valid path
in the search space of the SP algorithm. 

Now we assume this proposition is true of for $n = k, k \ge 1$ 
and prove the case when $n = k+1$. For a valid 
path $p^0 = (a_1, \cdots, a_k, a_{k+1} ) $, by our induction 
hypothesis, we can rearrange the first $k$ actions 
to obtain a path $(a_1^1, a_2^1, \cdots, a_k^1) $. 

Now we consider a new path $p^1 = (a_1^1, \cdots, a_k^1, a_{k+1})$.
There are two cases. First, if $L(a_{k+1}) < L(a_k^1)$,  or  
$L(a_{k+1}) > L(a_k^1)$  and $a_{k+1}$ is a follow-up action of $a_k^1$, 
then $p^1$ is already an SP-path. Otherwise, we 
have 
$L(a_{k+1}) > L(a_k^1)$ and $a_{k+1}$ is not a follow-up action of $a_k^1$. 
In this case, by Lemma~\ref{sp.left}, path $p^{1'}  = (a_1^1, \cdots, a_{k-1}^1, a_{k+1}, a_k^1$
is also a valid path that leads $s$ to the same state as $p_a$ does.

By the induction hypothesis, if $p^{1'}$ is still not an SP-path, we can rearrange 
the first $k$ actions in $p^{1'}$ to get a new path $p^2 = (a_1^2, \cdots, a_k^2, a_k^1)$. 
Otherwise we let $p^2 = p^{1'}$.
Comparing $p^1$ and $p^2$, we know $L(a_{k+1}) > L(a_k^1) $, namely, the level 
value of the last action in $p^1$ is strictly larger than that in $p^2$. We can repeat 
the above process to generate $p^3, \cdots, p^m, \cdots$ as long as $p^j (j \in Z^{+}) $ is not 
an SP-path. Our transformation from $p^j$ to $p^{j+1}$ also ensures that every $p^j$  is 
a valid path from $s$ and leads to the same state that $p_a$ does.

Since we know that the layer value of the last action in each $p_j$ is monotonically decreasing
as $j$ increases, such a process must stop after a finite number of iterations. Suppose 
it finally stops at $p^m = (a_1', a_2', \cdots, a_k', a_{k+1}'$, we must have that $L(a_{k+1}') \le L(a_k')$
or $L(a_{k+1}') > L(a_k')$ and $a_{k+1}'$ is a follow-up action of $a_{k'}$. Hence, 
$p^m$ now is an SP-path. We then assign $p^m$ to $p_b$ and the induction step is proved. 
\endproof

Theorem~\ref{thm.sp2} shows that the SP algorithm cannot reduce 
the number of states expanded in the search space. The reason is as follows: 
for any state in the original search space that is reachable from the initial 
state $s_0$ via a path $p$, there is still an SP-path that reaches $s$.
Therefore, every reachable state in the search space is still reachable 
by the SP algorithm. % However, 
In other words, SP reduces the number of generated states, but not the number of expanded states. 

SP is not a stubborn set based reduction algorithm.  This can be illustrated by the following example. 

Assuming a SAS+ planning task $S$ that contains two state variables $x_1$ and $x_2$, 
where both $x_1$ and $x_2$ have domain $\{0, 1\}$, with the initial state 
as $\{ x_1 = 0, x_2 = 0 \}$ and the goal as $\{ x_1 = 1, x_2 = 1\}$. Actions $a$
and $b$ are two actions in $S$ where $pre(a)$  is $ \{ x_1 = 0 \} $ and \textit{eff}$(a)$ is 
$\{x_1 = 1 \}$ and  $pre(b)$  is $\{ x_2 = 0 \} $ and \textit{eff}$(b)$ is 
$\{x_2 = 1 \}$. It is easy to see that $a$ and $b$ are not 
follow-up actions of each other, and that $x_1, x_2$ will be 
in different layers after stratification.  Without loss of generality, we can assume $L(a) = L(x_1) > L(x_2) = L(b)$. Therefore, we know that action $b$
will not be expanded after action $a$ in state $s: \{ x_1 = 1, x_2 = 0\}$. 
However, $apply(s, b)$ is the goal. Not expanding $b$ in state $s$ violates condition $A2$ 
in Definition~\ref{def.stubborn} where any valid path from $s$ to a 
goal state has to contain at least one action in the expansion set of $s$. 

We can also see in the above example that the search space explored by SP
contains four states, namely, the initial state $s_0$, $apply(s_0, a)$, $apply(s_0, b)$
and the goal state. Meanwhile, under the EC algorithm, in state $s_0$, the DTGs for 
$x_1$ and $x_2$ are not in each other's dependency closures.
This implies that in $s_0$, EC expands either action $a$ or $b$, but not both. 
Therefore, EC expands three states while SP expands four. 
This illustrates our conclusion
in Theorem~\ref{thm.sp2} that the SP algorithm cannot reduce 
the number of expanded states.

\section{A New POR Algorithm Framework for Planning}
\label{sec:algorithm}

We have developed a POR theory for planning and explained 
two previous POR algorithms using the theory. Now, based on the theory, we propose
a new POR algorithm which is stronger 
than the previous EC algorithm.

Our theory shows in Theorem~\ref{theo.left2} that the condition 
for enabling POR reduction is strongly related to left commutativity
of actions. In fact, constructing a stubborn set can be 
reduced to finding a left commutativity set. As we show in Theorem~\ref{theo.p2}, 
the EC algorithm follows this idea. However, the basic unit 
of reduction in EC is DTG (i.e. either all actions in a DTG are expanded or none of them are), which 
is not necessary according to our theory. Based on this insight, 
we propose a new algorithm that operates with the granularity of actions instead 
of DTGs.

\begin{defn} 
\label{defn:las}
For a state $s$, an action set $L$ is a
\textbf{landmark action set} if and only if any valid path starting from $s$ to a
goal state contains at least one action in $L$.
\end{defn}
 
\begin{defn} % support action
For a SAS+  task, an action $a \in \mathcal{O}$ is \textbf{supported} by an action $b$
if and only if $pre(a) \cap \textit{eff}(b) \neq \emptyset$.
\end{defn}

\begin{defn}    % This definition is state specific
\label{defn:adg}
For a state $s$, its \textbf{action support graph (ASG)} at $s$ is defined as a directed graph 
in which each vertex is an action, and there is an edge from $a$ to $b$ if and only if $a$ is not 
applicable in $s$ and $a$ is supported by $b$.
\end{defn}

The above definition of ASG is a direct extension of the definition of a causal graph. Instead of having domains as basic units, here we directly use actions as basic units. 

\begin{defn}
\label{defn:core}
For an action $a$ and a state $s$, the \textbf{action core} of $a$ at $s$, denoted 
by $AC_s(a)$, is the set of actions that are in the transitive 
closure of $a$ in $ASG(s)$. The action core for a given set of actions $A$ is the union 
of action cores of every action in $A$.
\end{defn}

\begin{lemma}
\label{lemma:ac}
For a state $s$, if an action $a$ is not
applicable in $s$ and there is a valid path $p$ starting from $s$ whose last action is $a$, 
then  $p$ contains an action $b, b \neq a, b \in AC_s(a)$.
\end{lemma}

\proof
We prove this by induction on the length of $p$.

In the base case where $|p| = 2$, we assume $p = (b,a)$. 
Since $a$ is not applicable in $s$, it must be supported
by $b$. Thus, $b \in AC_s(a)$.
Suppose this lemma is true for $ 2 \le |p| \le k-1$, we prove the case for $|p|=k$.
For a valid path $p=(o_1,\ldots, o_k)$, again there exists an action $b$ before $a$
that supports $a$. If $b$ is applicable in $s$, then $b \in AC_s(a)$. Otherwise,
we have a path $p' = (o_1, \ldots, b)$ with $2 \le |p'| \le k-1$. Thus,
by the induction assumption, $p'$ contains at least one action in $AC_s(b)$, which 
is a subset of $AC_s(a)$, according to Definition~\ref{defn:adg} and~\ref{defn:core}.
\endproof

\begin{defn}
\label{defn:commcol}  
Given a SAS+ planning task $\Pi$ with $O$ as set of all actions $O$,
for a state $s$ and a set of action $A$, the \textbf{action closure} of action set $A$ at $s$, denoted by 
by $C_s(A)$, is a subset of $O$ and a super set
of $A$ such that for any applicable action $a \in C_s(A)$ at $s$ and any action $b \in O 
\backslash C_s(A)$, $\textit{eff}(a)$ and $\textit{eff}(b)$ are conflict-free. In addition, 
if $pre(b) \in S$, $\textit{eff}(a)$ and $pre(b)$ are conflict-free. 
\end{defn}

Intuitively, actions in $C_s(A)$ can be executed without affecting the completeness and optimality of search. 
Specifically, because any applicable action in $C_s(A)$ and any action not in $C_s(A)$ will not assign different values to the same
state variable, for action $a \in C_s(A)$ and action $b \in O \backslash C_s(A)$ at $s$, path $(a, b)$ will lead to the same state that $(b, a)$ does.  Additionally, because $pre(b)$ and $eff(a)$ are conflict-free 
when $pre(b) \in s$, executing action $a$ will not affect the applicability of action $b$ in future. Therefore, 
actions in $C_s(A)$ can be safely expanded first during the search, while actions outside it can be expanded later. 
     
A simple procedure, shown in Algorithm~\ref{algo:ac}, can be used to find the action closure for a given action set $A$.

\begin{algorithm}[htp]
\SetKwInOut{Input}{input}
\SetKwInOut{Output}{output}
\Input{A SAS+ task with action set $O$, an action set $A \subseteq Q$, and a state $s$}
\Output{An action closure $C(A)$ of $A$}
\BlankLine
$C(A) \leftarrow A$\;
\Repeat{ $C(A)$ is not changing } {
    \ForEach{action $a$ in $C(A)$ applicable in $s$}
    {
        \ForEach{action $b$ in $O \backslash C(A)$} {
            \If{ $pre(b) \cap s \neq \emptyset$ 
            \textbf{and} $pre(b)$ and $\textit{eff}(a)$ are not conflict-free}
            {
                $C(A) \leftarrow C(A) \cup \{ b \}$ \;
            }
            \If{ $\textit{eff}(b)$ and $\textit{eff}(a)$ are not conflict-free } {
                $C(A) \leftarrow C(A) \cup \{ b \}$ \;
            }
        }
    }
}

\Return $C(A)$ \;
\caption{ A procedure to find action closure} \label{algo:ac}
\end{algorithm}

The proposed POR algorithm, called stubborn action core (SAC), works as follows. At any given 
state $s$, the expansion set $E(s)$ of state $s$ is determined by Algorithm~\ref{algo:sac}.

\begin{algorithm}
\SetKwInOut{Input}{input}
\SetKwInOut{Output}{output}
\Input{A SAS+ planning task and state $s$}
\Output{The expansion set $E(s)$}
\BlankLine

Find a landmark action set $L$ at $s$ \;
Calculate the action core $AC_s(L)$ of $L$ using Algorithm 1\;
Use $AC_s(L)$ as $E(s)$ \;

\caption{ The SAC algorithm} \label{algo:sac}
\end{algorithm}

There are various ways to find a landmark action set for a given state.
Here we give one example that is used in
our current implementation. To find a landmark action set $L$ at $s$, we utilize the DTGs associated with the SAS+ formalism. We first find a transition set that includes all possible transitions $(s_i, v_i)$ in an unachieved goal-related DTG $G_i$
where $s_i$ is the current state of $G_i$ in $s$. It is easy to see 
that all actions that mark transitions in this set make up a landmark action set, because $G_i$ is unachieved and at least one action starting from $s_i$ has to be performed in any solution plan. 

There are also other ways to find a landmark action set. For instance, the pre-processor 
in the LAMA planner~\cite{Richter08} can be used to find landmark facts, and all actions that lead to these landmark facts also make up a landmark action set. 

\begin{theorem}
For a state $s$, the expansion set $E(s)$ defined by the SAC algorithm is a stubborn set at $s$.
\end{theorem}
 
\proof

We first prove that our expansion set $E(s)$ satisfies condition 
$A1$ in Definition~\ref{def.stubborn}, namely, for any action 
$b \in E(S)$, and actions $b_1, \cdots, b_k \notin E(s)$, 
if $(b_1, \cdots, b_k, b)$ is a valid path from $s$, then
$(b, b_1, \cdots, b_k)$ is also a valid path, and 
leads to the same state that $ (b_1, \cdots, b_k, b)$ does. 

To simplify this proof, we can treat action 
sequence $(b_1, \cdots, b_k)$ as a ``macro'' action
$B$ where an assignment $x_t = v_t$ in $pre(B)$ if and only if
$x_t = v_t$ is in the precondition of some $b_i \in B$ and 
$x_t = v_t$ is not in the effects of a previous action $b_j (j < i)$, 
    and an assignment 
$x_t = v_t$ is in $\textit{eff}(B)$ if and only if 
$x_t = v_t$ is in the effect set of some $b_i \in B$, and
$x_t$ is not assigned to any value other than $v_t$ in the effects of later action $b_j (j > i)$. In the following proof, we use the macro action $B$ in place of the path $(b_1, \cdots, b_k) $.

To prove $A1$, we only need to prove that if $(B, b)$ is a valid path, then 
$s: b\Rightarrow B$. According to Theorem~\ref{theo.comm}, 
$s: b\Rightarrow B$   if and only if the following four propositions are true.
    
a) Action $b$ must be applicable in $s$. We prove this by contradiction. Let $s' = apply(s, B)$, 
if $b$ is not applicable in $s$, 
but applicable in $s'$, then
$B$ supports $b$. Since all effects of $B$ are from 
actions in the path $(b_1, \cdots, b_k)$, there exists
an action $b_i \in \{b_1, \cdots, b_k\}$ such that $b_i$ supports $b$. However,
according to Definition~\ref{defn:core}, $b_i$ is in the transitive closure
of $b$ in $ASG(s)$. According to our algorithm, $b_i$ should be in $E(s)$. This 
contradicts with our assumption that $b_i \notin E(s)$. Thus, $b$ must be applicable 
at $s$. 

b) $pre(B)$ and $\textit{eff}(b)$ are conflict-free. We prove this proposition by contradiction. 
If $pre(B)$ and $\textit{eff}(b)$ are not conflict-free, we assume that 
$pre(B)$ has $x_t =v_t$ that conflicts with an assignment in $\textit{eff}(b)$. According 
to the way we define $B$, there exists an action $b_i \in (b_1, \cdots, b_k)$, 
such that $x_t = v_t$. 
Also, since $B$ is applicable in $s$, we know that $x_t$ takes the value $v_t$
at $s$ also. Therefore, we know that $pre(b_i)$ and $\textit{eff}(b)$ are not conflict-free.
However, according to Definition~\ref{defn:commcol} and Algorithm~\ref{algo:ac}, $b_i$
is in $E(s)$. This contradicts with our assumption that $b_i$ is not in $E(s)$. 
Thus, $pre(B)$ and $\textit{eff}(b)$ are conflict-free. 

c) \textit{eff}$(B)$ and $\textit{eff}(b)$ are conflict-free. The proof of this proposition is 
very similar to the one above. If they are not conflict-free, we must 
have action $b_i \in (b_1, \cdots, b_k)$, such that $\textit{eff}(b)$ and $\textit{eff}(b_i)$
are not conflict-free. However, according to Definition~\ref{defn:commcol} and 
Algorithm~\ref{algo:ac}, $b_i$ is in $E(s)$. This contradicts with our assumption that 
$b_i$ is not in $E(s)$.  Thus, $\textit{eff}(B)$ and $\textit{eff}(b)$ are conflict-free. 

d) $pre(b)$ and $\textit{eff}(B)$ are conflict-free. This proposition is true 
as we assumed in condition $A1$ that $(B, b)$ is a valid path from $s$.

Thus, from Theorem~\ref{theo.comm}, we see that $s: b \Rightarrow B$ and that 
condition $A1$ in Definition~\ref{def.stubborn} is true. 

Now we verify condition $A2$ by showing that any solution path $p$ from $s$ contains at least 
one action in $E(s)$.  From the definition of landmark action sets, we know that there 
exists an action $l \in L$ such that $p$ contains $l$. From Lemma~\ref{lemma:ac} we know 
that $AC_s(l)$ contains at least one action, applicable in $s$, in $p$. Thus, $E(s)$ indeed 
contains at least one action in $p$.  

Since $E(s)$ satisfies conditions $A1$ and $A2$ 
in Definition~\ref{def.stubborn}, $E(s)$ is a stubborn set in state $s$.  

\endproof

% An example
\subsection{SAC vs. EC}

\begin{figure}[t]
\centering
\ifpdf
	\includegraphics[scale=0.2]{a.png}
\else
	\includegraphics[scale=0.2]{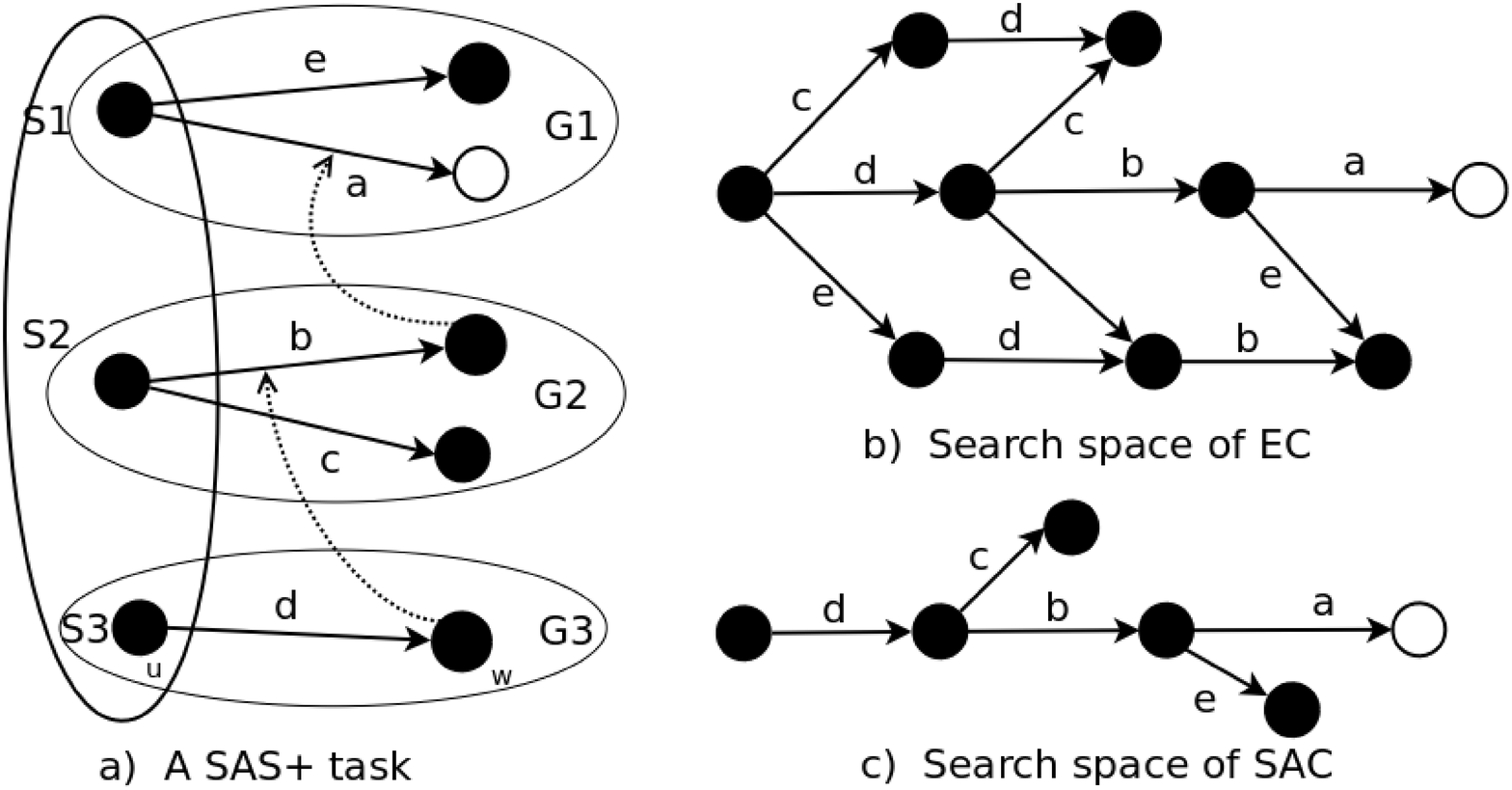}
\fi	
\caption{\label{ac-ec} Search spaces of EC and SAC}
\end{figure}

SAC gives stronger reduction than the previous EC algorithm, since it 
is based on actions, which have a finer granularity than DTGs do.
Specifically, SAC gives more reduction than EC for two reasons.
First, applicable actions that are not associated with landmark
transitions, even if they are in the same DTG,
are expanded by EC but not by SAC. 
Second, applicable actions that do not support
any actions in the landmark action set, even if they are in the same DTG,
are expanded by EC but not by SAC.  

To give an example, in Figure~\ref{ac-ec}a, G1,
G2, G3 are three DTGs. The goal assignment is marked as an unfilled circle in G1.
$a, b, c, d, e$ are actions. Dashed arrows denote the preconditions of actions. For
instance, the lower dashed arrow means that $b$ requires a 
precondition $x_3 = w$.
 
In this example, according to EC,
G1 is a goal DTG and G2 and G3 are in the dependency
closure of G1. Thus, before executing $a$,
EC expands every applicable action in G1, G2 and G3
at any state. 
SAC, on the other hand,
starts with a singleton set $\{a\}$ as the initial landmark
action set and ignores action $e$. Applicable action $c$ is also not included
in the action closure in state $s$ since it does not support $a$. The
search graphs are compared in Figure~\ref{ac-ec} and we see 
that SAC gives stronger reduction.

\section{System Implementation}

\begin{figure}[t]
\centering
\ifpdf
	\includegraphics[scale=0.65]{fdsac.png}
\else
	\includegraphics[scale=0.65]{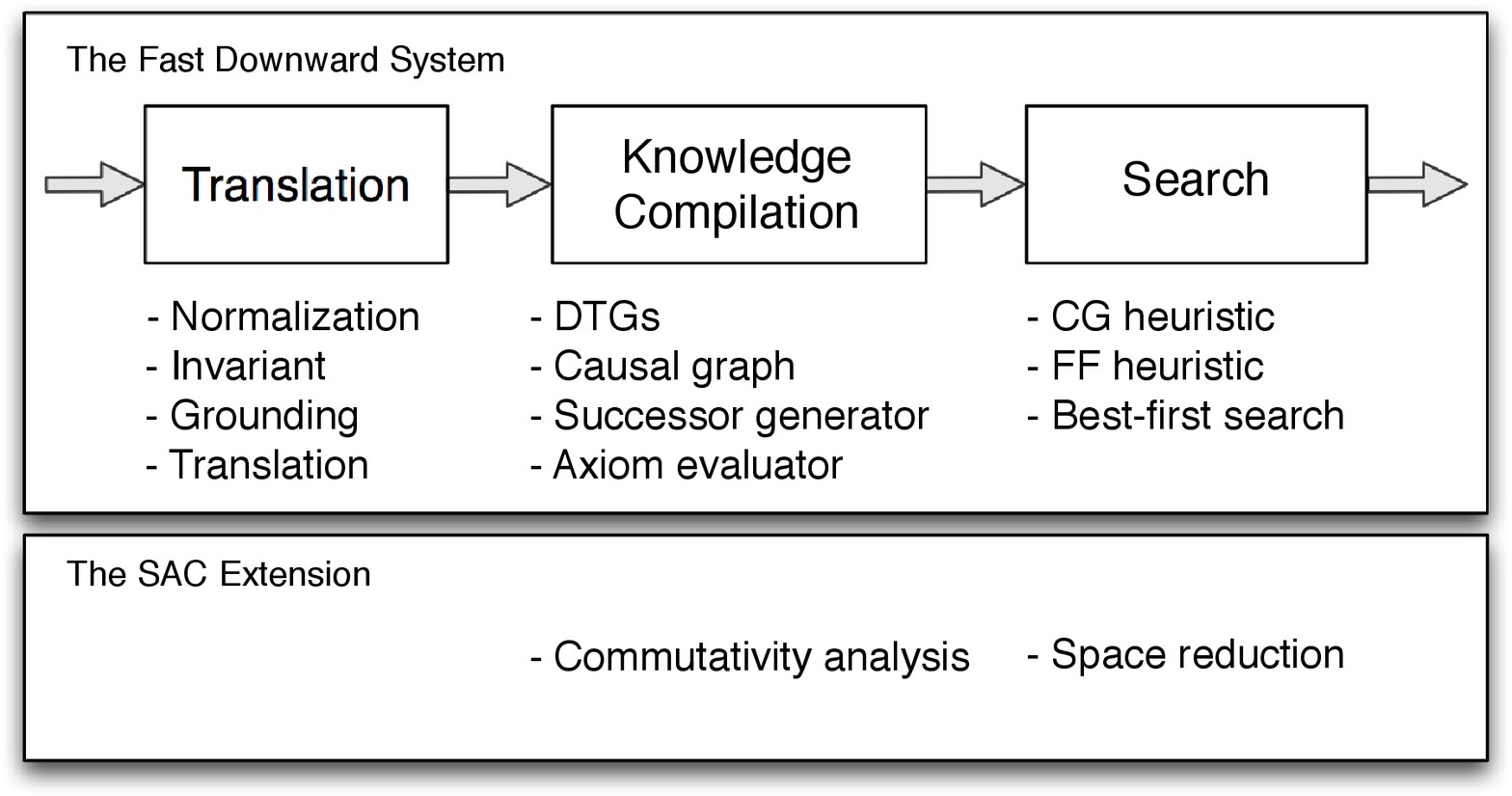}
\fi	
\caption{\label{fdsac} System Architecture of FD and SAC}
\end{figure}

We adopt the Fast Downward (FD) planning system~\cite{helmert06} as our code base. The 
overall architecture of FD is described in Figure~\ref{fdsac}. 
A complete FD system contains three parts 
corresponding to three phases in execution: translation, knowledge
compilation and search. Translation module will 
convert planning tasks described into 
a SAS+ planning task. The knowledge compilation 
module will generate domain transition graphs and causal 
graph for the SAS+ planning task. The search module
implements various state-space-search algorithms as 
well as heuristic functions. All these three modules
communicate by temporary files. 

We make two additions to the above system to implement our SAC planning system, as shown 
in Figure~\ref{fdsac}. 
First, we add a ``commutativity analysis'' module into the knowledge compilation step 
to identify commutativity between actions.  Second, we add a ``space reduction'' module to the search module to conduct state space reduction.  
The commutativity analysis module is used to build left commutativity relations
between actions and build the action support graph. It reads 
action information from the output of knowledge compilation 
module and determines the left commutativity relations between actions 
according to conditions in Theorem~\ref{theo.left2}. In addition, 
this module also determines if one action is supported by another and builds the action support graph defined in Definition~\ref{defn:adg}.  
The reduction module for search is used to generate a stubborn set of a given state. 
We implement the SAC algorithm in this module. Starting from a landmark action set $L$ as
the target action set, we find the action closure $AC_s(L) $ iteratively add actions that support
actions in the target action set to the target action set until it is not changing. 
We then use the applicable actions in the action closure as the 
set of actions to expand at $s$.  In other words, in our SAC system, 
during the search, for any given state $s$, instead of using 
successor generator provided by FD to generate a set of applicable 
operators, we use the reduction module to generate a stubborn set in state $s$ and 
use it as the expansion set. 

It is easy to see that the overall time complexity of determining 
left commutativity relationships between actions is $O(|A|^2)$ where 
$|A|$ is the number of actions.  We implement this module in Python. 
Since the number of actions $|A|$ is usually not large, in most of the cases, 
the commutativity analysis module takes less than 1 second to finish. 
This module only runs once for solving a planning problem. Therefore, 
the commutativity analysis module amounts to an insignificant amount 
of overhead to the system.  Theoretically, the worst case time complexity for finding the action closure 
is $O(|A|^2)$ where $|A|$ is the number of actions. However, in practice, by 
choosing the landmark action set $L$ that associated with 
transitions in an unarchived goal-related DTG starting from 
current state, the procedure of finding action closure terminates 
quickly after about 4 to 5 iterations.  Therefore, adding 
the reduction module does not increase the overall search overhead
significantly either. We implement this module in C++
and incorporate it into the search module of FD. 

\section{Experimental Results}
\label{sec:results}

We test our algorithm on problems in the recent International Planning
Competitions (IPCs): IPC4, and IPC5. We implemented our algorithm
on top of the Fast Downward (FD) planner~\cite{helmert06}. 
We only modified the state expansion part.

We have implemented our SAC algorithm and tested it along with Fast Downward and its combination 
with the EC extension on a Red Hat Linux server with 2Gb memory and one 2.0GHz CPU. The admissible HSP $h_{max}$ heuristic~\cite{Bonet01}
and inadmissible Fast Forward (FF) heuristic~\cite{Hoffmann01} are used in our experiments. 

First, we apply our SAC algorithm to $A^*$ search with the HSP $h_{max}$ heuristic~\cite{Bonet01}. We also 
turn off the option of preferred operators~\cite{helmert06} since it compromises the optimality of $A^*$ search. 
Table~\ref{tbl.astar} shows the detailed results on node expansion and generation during the search. 
We also compare the solving times of these three algorithms. As we can clearly see
from Table~\ref{tbl.astar}, the numbers of expanded nodes of the SAC-enhanced $A^*$ algorithm are consistently lower than those of the baseline $A^*$ algorithm and the EC-enhanced $A^*$ algorithm. There are some cases where the generated nodes of the SAC-enhanced algorithm are slightly larger than those of the baseline $A^*$ or EC-enhanced $A^*$ algorithm. This is possible due to the tie-breaking of states 
with equal heuristic values during search. 

We can also see that the computational overhead of SAC is low. For instance, in the Freecell domain, the running time of the SAC-enhanced algorithm is only slightly higher than the baseline and lower than the EC-enhanced algorithm, despite their equal number of expanded and generated nodes. 

Aside from the $A^*$ algorithm, we also test SAC on best-first search algorithms.  
Although POR preserves completeness and optimality, it can also be combined 
with suboptimal searches such as best-first search 
to reduce their search space. In this comparison, 
we turned off the option of preferred operators in our experiment for FD. 
Preferred operator is another space reduction method that 
does not preserve completeness, and using it with EC or SAC will 
lead to worse performance. We will investigate 
how to find synergy between these two approaches in our future work. 
We summarize the performance of three algorithms, original Fast Downward (FD), FD with EC, and FD with 
SAC, in Table~\ref{tbl.2} by presenting 
the number of problem instances in a planning domain that can be solved within 1800 seconds by each solver. 
We also ignore small problem instances with solving time less than $0.01$ seconds. 
All there solvers uses inadmissible Fast Forward (FF) heuristic. 
As we can see from Table~\ref{tbl.2}, when combined with a best-first-search algorithm, SAC can still reduce the number of generated and expanded nodes
compared to the baseline FD algorithm and the EC-enhanced algorithm. 
In many problems (e.g. {\em pipesworld18}, {\em tpp15}, {\em truck13}), the saving on the number of expanded 
states can be of orders of magnitude.

\section{Related Work}
\label{sec:related}

We discuss some related work in this section.  

\subsection{Symmetry}

Symmetry detection is another way for reducing the search space~\cite{Fox99}. From the view of
node expansion on a SAS+ formalism for planning, we can see that
symmetry removal is different from SAC. For example, consider a
domain with three objects $A_1$, $A_2$, and $B$, where $A_1$ and
$A_2$ are symmetric, and actions associated with $B$ have
no conflict with any actions associated with $A_1$ or
$A_2$. In this case, symmetry removal will expand actions associated with ($A_1$
and $B$) or ($A_2$ and $B$), whereas SAC will only expand actions 
associated with the DTG for $B$ if we pick a landmark action 
set based on it. This is because both the action core and the action closure 
will not include any actions associated with $A_1$ or $A_2$ as
they have no conflict with actions in $B$.

Intuitively, symmetric removal finds that it is not important
whether $A_1$ or $A_2$ is used since they are symmetric, whereas
SAC finds that it is not important whether actions associated with $B$ is used before or
after actions associated with $A_i, i=1,2$ since there is no conflict.
In fact, SAC can also detect
stronger relationships such as the fact that it is safe to use actions associated with $B$ before
those associated with $A_i$, since any path that uses actions associated with $A_i$ before those associated with $B$ corresponds to
another valid path with the same cost.

Further, there is limited research on domain-independent detection
and removal of symmetry. The method by Fox and Long~\cite{Fox99}
detects symmetry from the specification of initial and goal states and may miss many symmetries.

\subsection{Factored planning}

Factored planning~\cite{Amir03,Bradman06,Kelareva07} is a class of
search algorithms that exploits the decomposition of state space.
In essence, factored planning finds all the subplans for each
individual subgraph and tries to merge them. There are some
limitations of factored planning.

First, for some problems with dense subgraphs, the number of
subplans in each subgraph may be very large, making the search
very expensive. What is worse is that there are many subgraphs in
which the goal is not specified, leading to more subplans that
need to be considered. We have done some empirical study on this
matter. For example, for pipesworld20, there are 96 DTGs, 18 out
of which have goal facts. Even if we only consider one DTG and
apply the \emph{canonicality assumption} that each state can be
included at most once in any subplan, the number of subplans from
the initial state to the goal can be as high as 1.96$\times 10^{9}$ in DTG
\#16 generated by Fast Downward. The number is high because there
are multiple transition paths, and each transition can be
associated with many actions. If we multiply the numbers of
possible subplans of the 18 DTGs containing goals, the number will
approximately be of the order of $10^{120}$. Thus, the search space
will be extremely large if we consider all the 96 DTGs (78 of
which do not even have a goal state) and remove the canonicality
assumption. Of course, techniques such as tree search and
pruning~\cite{Kelareva07} can speed up the process but the
potential speedup is largely unknown.

Second, since the canonicality assumption is generally not true
for many domains, and there are potentially infinite number of
subplans without restriction on the subplan length, the factored
planning algorithm needs to use certain schemes such as iterative
deepening~\cite{Bradman06} to restrict the subplan length. These
schemes further increase the complexity and may compromise the
global optimality of the resulting plan~\cite{Kelareva07}.

In summary, although factored planning has shown potential on some
domain-dependent studies, its practicality for general
domain-independent planning has not been established yet. We note
that POR algorithms we studied in this paper
are not exclusive to factored planning and it is possible
that POR can be integrated into factored planning
to reduce the cost of search.

\subsection{Planning utilizing the SAS+ decomposition}

Our POR method is based on the SAS+ representation~\cite{helmert06}. Recently
there has been increasing interest in utilizing the SAS+
representation.

The Fast Downward planner~\cite{helmert06} develops its
heuristic function by analyzing the causal graphs on top of the
SAS+ models. Another SAS+ based heuristic for optimal planning is
recently obtained via a linear programming model encoding
DTGs~\cite{CP07}. The LAMA planner derives inadmissible heuristic values by analyzing landmarks in SAS+ models~\cite{Richter08}. An admissible version of it is proposed in~\cite{Karpas09} by using action cost partitioning. 
Yet another admissible heuristic called `merge-and-shrink' is developed
based on abstraction of domain
transitions~\cite{helmert:icaps-2007}, which strictly
dominates the admissible landmark heuristics~\cite{HelmertD09}. 
Moreover, long-distance
mutual exclusion constraints based on a DTG analysis is proposed
and shown to be effective in speeding up SAT-based optimal
planners~\cite{AIJ08}. The DTG-Plan planner searches
directly on the space of DTGs in a hierarchical decomposition
fashion~\cite{Chen:AAAI-08}. The algorithm is shown to be fast but
is not complete or optimal. 

Comparing to the above recent work, POR offers a
completely new approach to exploit the state-space decomposition
in the SAS+ representation. It is orthogonal to the design of
better heuristics and it provides a systematical, theoretically sound way
to reduce search costs.

POR is most effective for problems where the action support graphs are directional and the inter-action dependencies are not dense. It may not be useful for problems where the actions are strongly connected \emph{and}
there is a high degree of inter-action dependencies. For example, it
is not useful for the 15-puzzle where each action 
on each piece is supported 
by surrounding actions, which makes 
the action support graph strongly connected. 
In this case, POR cannot give reductions during the search.

\section{Conclusions and Future Work}
\label{sec:conclusion}

Previous work in both model checking and AI planning 
has demonstrated that POR 
is a powerful method for reducing
search costs. POR is an {\em enabling} technique for modeling checking, which 
will not be practical without POR due to its high complexity. Although 
POR has been extensively studied for model checking, its theory has not been developed for AI planning. 
In this paper, we developed a new POR theory for planning 
that is parallel to the stubborn set theory 
in model checking.

In addition, by analyzing the structure of actions in planning problems,
we derived a practical criterion that 
defines left commutativity between actions. 
Based on the notion of left commutativity, 
we developed sufficient conditions 
for finding stubborn sets during search for planning. 

Furthermore, we applied our theory to explain two previous POR algorithms for planning.  
The explanation provided useful insights that lead to a stronger and more efficient 
POR algorithm called SAC.  Compared to previous POR algorithms, SAC finds stubborn sets based on a finer granularity for checking left commutativity, leading to strong reduction. We compared the performance of SAC to 
the previously proposed EC algorithm on both optimal and non-optimal state space searches. Experimental results showed 
that the proposed SAC algorithm led to significantly stronger node reduction and less overhead. 

In our future work, we plan to develop stronger POR algorithms for
planning based on our theoretical framework and study its interaction
with other search reduction techniques such as preferred operators~\cite{helmert06}, abstraction heuristics~\cite{helmert:icaps-2007}, landmarks~\cite{Richter08}, and symmetry
detection~\cite{Fox99}.

%------------------------

\bibliographystyle{acmtrans}
\bibliography{Chen}

\begin{landscape}
\clearpage
\setlength{\LTcapwidth}{\textwidth}
\setlongtables

\begin{longtable}{|c|ccc|ccc|ccc|}
\caption{\small Comparison of FD, EC, and SAC using A*
with $h_{max}$ heuristic on IPC domains.  We show numbers of expanded and generated nodes. ``-'' means timeout after 300 seconds. 
For each problem, we also highlight the best values of expanded and generated nodes among three algorithms, if there is any difference.
} \label{tbl.astar}\\
\hline\hline
Domain  &  \multicolumn{3}{|c|}{FD} &  \multicolumn{3}{|c|}{EC} & \multicolumn{3}{|c|}{SAC} \\ 
\cline{2-10}
        &  Expanded    & Generated  &    Time     & Expanded  &    Generated  & Time & Expanded  &    Generated  & Time\\ \hline
\endfirsthead

\multicolumn{10}{l}{continued from previous page} \\ \hline
  Domain  &  \multicolumn{3}{|c|}{FD} &  \multicolumn{3}{|c|}{EC} & \multicolumn{3}{|c|}{SAC} \\
  \cline{2-10} 
  &  Expanded    & Generated  &    Time     & Expanded  &    Generated  & Time & Expanded  &    Generated  & Time \\\hline
\endhead
 
\hline
 \multicolumn{10}{r}{{continued on next page}} \\
\endfoot
\hline
\endlastfoot

$ airport1 $ & 9 & 9 & 0 & 9 & 9 & 0 & 9 & 9 & 0 \\
$ airport2 $ & 16 & 17 & 0 & 16  & 17 & 0 & 16 & 17 & 0 \\
$ airport3 $ & 38 & 102 & 0 & 35 & 90 & 0.01 & \ovalbox{27} & \ovalbox{43} & 0.01 \\
$ airport4 $ & 21 & 21 & 0 & 21 & 21 & 0.01 & 21 & \ovalbox{20} & 0.01\\
$ airport5 $ & 22 & 28 & 0 & 22 & 28 & 0.01 & 22 & \ovalbox{23} & 0.01 \\
$ airport6 $ & 138 & 335 & 0.01 & 120 & 230 & 0.07 & \ovalbox{86} & \ovalbox{202} & 0.06 \\
$ airport7 $ & 221987 & 5305641 &  81.02  & 221987 & 5305641 &  91.07   & \ovalbox{221901} & \ovalbox{709575} & 74.95  \\
$ airport8 $ & 2420 & 11190 & 0.73& 2364 & 6621 & 1.93 & \ovalbox{699} & \ovalbox{2860} & 0.56 \\
$ airport9 $ &11005&60058 & 4.66 & 11005 & 39027 &16.35& \ovalbox{4923}& \ovalbox{24244} & 5.86 \\
$ airport10 $ & 19 & 20 & 0.01 & 19 & 20 & 0.01 & 19 & \ovalbox{18} & 0.01 \\
$ airport11 $ & 22 & 28 & 0.01 & 22 & 28 & 0.01 & 22 & \ovalbox{23} & 0.01 \\
$ airport12 $ & 122 & 328 & 0.03 & 104 & 119 & 0.07 & \ovalbox{80} & \ovalbox{195} & 0.06 \\
$ airport13 $ & 112 & 295 & 0.01 & 94 & 182 & 0.07 & \ovalbox{76} & \ovalbox{187} & 0.08 \\
$ airport14 $ & 2300 & 11144 &0.84 & 2246 & 6324 & 2.14 & \ovalbox{626} & \ovalbox{2700} & 0.61 \\
$ airport15 $ & 1910 & 9240 & 0.69 & 1904 & 5396 & 1.94 & \ovalbox{493} & \ovalbox{2124} & 0.46 \\
\hline
$ depot1 $ & 159 &  1000&   0.01&   159&    1000&   0.02&   159&    \ovalbox{981}&0.02 \\
$ depot2 $ & 2294&   17803&  0.01&   2310&   17894&  0.52&   2294&   \ovalbox{16404} & 0.34 \\
$ depot3 $ &2389 &21172 & 0.2 & 2389 & \ovalbox{20724} & 0.3 & 2389 & 21168 & 0.76\\  %%
$ depot4 $ &42435& 366989 & 5.08 & 42435 & 366989 & 7.08 & \ovalbox{42362} & \ovalbox{364883} & 4.35 \\ %% 
$ depot5 $ & 13096 & 119388 & 2.35 & 13096 & 119388 & 4.15 & 13096 & 119388 & 3.07 \\ %%
$ depot7 $ & 9672 & 91795 & 0.81 & 9672 & 91795 & 1.89 &  \ovalbox{9658} & \ovalbox{91460} & 1.73 \\%%
$ depot8 $ & 173184 & 1999709 & 40.15 & 173184 & 1999709 & 60.8 & \ovalbox{173157} & \ovalbox{1993740} & 51.68 \\ %%

\hline
$ drivelog1 $ & 57 & 373 & 0 & 57 & 355 & 0 & \ovalbox{30} & \ovalbox{183} & 0 \\
$ drivelog2 $ & 55780 & 417679 & 2.35 & 55387 & 393334 & 3.2 & \ovalbox{52124} & \ovalbox{325004} & 4.12 \\
$ drivelog3 $ & 2982 & 22693 & 0.12 & 2858 & 21247 & 0.16 & \ovalbox{2682} & \ovalbox{19774} & 0.13 \\
$ drivelog4 $ & 460727 & 4798803 & 31.33 & 446341 & 4175538 & 35.38 & \ovalbox{414148}
& \ovalbox{4004775} & 28.65\\ %%
$ drivelog5 $ & 2077987 & 24224013 & 167.86 & 2040088 & \ovalbox{22120590} & 202.42&  \ovalbox{1943657} & 22156584 & 168.73 \\ %%

\hline
$ freecell1 $ & 63 & 407 & 0.01 & 63 & 407 & 0.01 & 63 & 407 & 0.03 \\
$ freecell2 $ & 212 & 1603 & 0.06 & 212 & 1603 & 0.18  & 212 & 1603 & 0.23 \\
$ freecell3 $ & 162 & 1156 & 0.07 & 162 & 1156 & 0.37 & 162 & 1156 & 0.26 \\
$ freecell4 $ & 792 & 4765 & 0.35 & 792 & 4765 & 2.35 & 792 & 4765 & 1.76 \\
$ freecell5 $ & 526 & 2930 & 0.49 & 526 & 2930 & 1.81 & 526 & 2930 & 1.81 \\
$ freecell6 $ & 430 & 2834 & 0.88 & 430 & 2834 & 3.39 & 430 & 2834 & 2.38 \\
$ freecell7 $ & 1429 & 8347 & 2.5 & 1429 & 8347 & 5.5 & 1429 & 8347 & 4.02 \\
$ freecell8 $ & 1682 & 12015 & 6.68 &  1682 & 12015 & 16.21 &  1682 & 12015 & 14.66 \\
$ freecell9 $ & 2001 & 12122 & 7.44  & 2001 & 12122 & 18.42& 2001 & 12122 & 16.32 \\
$ freecell10 $ & 1953 & 14383 & 11.66& 1953 & 14383 & 31.57 & 1953 & 14383 & 23.54 \\
\hline
$ rover1 $ &  323&    2043& 0 &   323&    2043&   0.01&   \ovalbox{114}&    \ovalbox{558} & 0\\
$ rover2 $ & 161 &    1019& 0 &  161&    1019& 0 &    \ovalbox{64}&     \ovalbox{239}& 0   \\
$ rover3 $ & 863&    5724&  0.01&   863&    5724&   0.04&   \ovalbox{390}&    \ovalbox{2007}&   0.01\\
$ rover4 $ & 291&    2399 &   0 & 291&    2399&   0.01&   \ovalbox{79}&     \ovalbox{397}& 0 \\
$ rover5 $ & 324204 & 5714884 & 6.42 & 324204 & 5714884 & 9.23 & \ovalbox{95369} &  \ovalbox{1093514} & 5.1 \\%%
$ rover6 $ & - & - & - & - & - & - & \ovalbox{251276} & \ovalbox{2437578} & 10.48 \\%%
$ rover7 $ & 156150 & 2179859 & 2.64& 156150 & 2179859 & 5.21 & \ovalbox{34772} & \ovalbox{387527} & 2.01 \\%%
\hline
$ truck1 $ & 390&    4306&   0.03&   390&    4306&   0.03&   390&    \ovalbox{1145}&   0.04 \\
$ truck2 $ & 943&    13212&   0.01&   943&    13212&  0.12&   \ovalbox{942}&    \ovalbox{2432}&   0.11 \\
$ truck3 $ & 9162&   178481& 1.12&   9162&   178481& 1.82&   \ovalbox{9157}&   \ovalbox{44098}&  1.03 \\
$ truck4 $ & 37799 & 258286 & 1.77 & 8792 & 15177 & 1.77 & \ovalbox{6841} & \ovalbox{12570} & 1.51 \\ %%
\hline \hline
\end{longtable}
\end{landscape}

\begin{landscape}
\clearpage
\setlength{\LTcapwidth}{\textwidth}
\setlongtables

\begin{longtable}{|c|ccc| ccc| ccc |}
\caption{\small Comparison of FD, EC and SAC with no-preferred operators on IPC's domains. 
We show numbers of expanded and generated nodes. ``-'' means timeout after 1800 seconds. For each problem, we also highlight the best values of expanded and generated nodes among three algorithms
  if there is any difference.}
\label{tbl.2} \\
\hline\hline
Domains & \multicolumn{3}{|c|}{FD} & \multicolumn{3}{|c|}{EC}  &\multicolumn{3}{|c|}{SAC}\\
\cline{2-10}
& Expanded & Generated & Time & Expanded & Generated & Time & Expanded & Generated & Time \\ \hline
\endfirsthead

\multicolumn{10}{l}{continued from previous page}  \\
\hline Domains & \multicolumn{3}{|c|}{FD} & \multicolumn{3}{|c|}{EC}  &\multicolumn{3}{|c|}{SAC}\\
\cline{2-10}
& Expanded & Generated & Time &Expanded & Generated & Time & Expanded & Generated & Time \\ \hline
\endhead

\hline
\multicolumn{10}{r}{{continued on next page}} \\
\endfoot

\hline
\endlastfoot
%DONE
$airport13 $  &\ovalbox{43}&	106&	0.03&	46&	102&	0.08&	45&	\ovalbox{68}&	0.05 \\
$airport14 $  &78&	321&	0.07&	70&	199&	0.19&	70&	\ovalbox{161}&	0.12 \\
$airport15 $  & 67&		224&	0.06&	66&	163&	0.2&	\ovalbox{64}&	\ovalbox{141}&	0.12 \\
$airport16 $  &  310&	1541&	0.35&	315&	1549&	1.05&	310&	\ovalbox{740}&	0.62   \\
$airport17 $  &  815&	4682&	1.15&	21819&	140496&	101.16&	\ovalbox{809}&	\ovalbox{2315}&	2.1   \\
$airport18 $  &  \ovalbox{18653}&	142281&	44.3&	82007&	655661&	529.57&	18712&	\ovalbox{76214}&	78.78    \\
$airport19 $  &  10564&	65299&	15.07&	10621&	65480&	48.12&	\ovalbox{5041}&	\ovalbox{17123}&	15.5   \\
$airport20 $  &	 \ovalbox{25377}&	182487&	52.94&	150816&	1272996&	959.32&	25636& \ovalbox{116148} &	53.58 \\
$airport21 $  &	 102&	274&	0.24&	102&	256&	1.57&	102&	\ovalbox{193}&	0.56 \\
$airport22 $  &	 149&	524&	0.51&	150&	509&	3.34&	149&	\ovalbox{370}&	1.52 \\
$airport23 $  &	 169&	620&	0.81&	169&	561&	6.03&	169&	\ovalbox{500}&	1.51	\\
$airport24 $  &	 166&	888&	1.14&	1902&	7459&	70.58&	\ovalbox{111}&	604&	1.08	\\
$airport25 $  &	 33751&	197109&	256.95&	-	&	-	& 	-	& 	\ovalbox{18752}&	\ovalbox{66186}	& 161.84	\\
%DONE
\hline
$driveslog11 $&24&    284&	0.01&	24&	284&	0.02&	87&	949&	0.05	\\
$driveslog12 $&78&	1059&	0.05&	\ovalbox{64}&	\ovalbox{224}&	0.04&	150&	1875&	0.13	\\
$driveslog13 $&583&	6396&	0.25&	75&	864&	0.06&	75&	864&	0.07	\\
$driveslog14 $&248&	3378&	0.13&	75&	883&	0.07&	75&	\ovalbox{416}&	0.05	\\
$driveslog16 $&  64971&	1331293&	146.32&	50855&	1017221&	110.91&	\ovalbox{38746}&	\ovalbox{737836}&	142.11    \\
$driveslog17 $&  24860&	804116&	204.9&	149938&	5012829&	1253.07&	\ovalbox{15620}&	\ovalbox{537050}&	138.37   \\
$driveslog19 $	& 834&	26406&	9.77&	727&	14443&	7.37&	\ovalbox{573}&	\ovalbox{1105}&	5.19  \\
% DONE
\hline
$freecell1     $  &10	&59&	0.08&	11&	63&	0.09&	10&	\ovalbox{54}&	0.08\\
$freecell2     $  &17&	114&	0.1&	17&	114&	0.1&	17&	\ovalbox{100}		& 0.01\\
$freecell3     $  &20&	144&	0.3&	\ovalbox{19}&	\ovalbox{122}&	0.14&	20&	127&	0.13\\
$freecell4     $  &48&	268&	0.39&	61&	401&	0.26&	\ovalbox{35}&	\ovalbox{238}&	0.17\\
$freecell5     $  &88&	510&	0.62&	123&	668&	0.82&	\ovalbox{63}&	\ovalbox{236}&	0.35\\
$freecell6     $  &   98&	510&	0.74&	97&	496&	0.71&	\ovalbox{58}&	\ovalbox{384}&	0.7	\\
$freecell7     $  &   412&	1594&	4.26&	233&	884&	1.93&	\ovalbox{215}&	\ovalbox{814}&	2.23   \\
$freecell8     $  &   106&	700&	1.02&	\ovalbox{71}&	\ovalbox{436}&	0.83&	85&	501&	1.13   \\
$freecell9     $  &   \ovalbox{68}&	\ovalbox{500}&	2.6&	112&	667&	3.93&	122&	719&	3.3 \\
$freecell10    $  &   797&	5043&	50.79&	4057&	34183&	217.53&	\ovalbox{112}&	\ovalbox{597}&	5.49  \\
$freecell11    $  &   120&	530&	3.72&	119&	620&	4.07&	\ovalbox{118}&	\ovalbox{527}&	4.21  \\
$freecell12    $  &   192&	1085&	5.6&	\ovalbox{58}&	\ovalbox{420}&	2.58&	96&	491&	3.51 \\
$freecell13    $  &   \ovalbox{412}&	\ovalbox{2655}&	10.21&	589&	2896&	9.26&	592&	2996&	13.16 \\
$freecell14    $  &   287&	2274&	31.27&	339&	2054&	29.96&	\ovalbox{78}&	\ovalbox{560}&	8.86  \\
$freecell15    $  &   1913&	8642&	140.72&	531&	2795&	35.44&	\ovalbox{248}&	\ovalbox{1197}&	18.21 \\
$freecell16    $  &   3818&	31173&	143.88&	409&	2381&	10.87&	\ovalbox{371}&	\ovalbox{1722}&	7.14 \\
$freecell17    $  &   246&	1402&	19.71&	244&	2849&	23.62&	\ovalbox{104}&	\ovalbox{740}&	10.99 \\
$freecell18    $  &	  3480&	\ovalbox{17263}&	354.43&	6069&	32509&	685.41&	\ovalbox{3036}&	19476&	174.69 \\
% DONE
\hline

$pipesworld7 $&   13&	685&	1.19&	13&	685&	0.73&	13&	685&	0.71   \\
$pipesworld8 $&   12&	550&	0.79&	12&	\ovalbox{473}&	0.74&	12&	531&	0.79  \\
$pipesworld9 $&  319&	8297&	10.05&	\ovalbox{165}&	5002&	6.31&	166&	\ovalbox{4996}&	5.74   \\
$pipesworld10$&  \ovalbox{50}&	\ovalbox{1572}&	2.92&	115&	3322&	5.14&	168&	4684&	7.4 \\
$pipesworld11$&  \ovalbox{197}&	\ovalbox{970}&	0.34&	451302&	2168501	&	459.77	&	1615&	7311&	4.24     \\
$pipesworld13$&   - &   - &   	-   &   8083 & 	54502 	& 	40.44   &   \ovalbox{4272}    &   \ovalbox{20342}   &   6.16 \\
$pipesworld14$&  3412&	16652&	6.01&	\ovalbox{2520}&	\ovalbox{11302}&	4.33&	2815&	13066&	10.66  \\
$pipesworld15$&  568191&	2792188&	1284.82&	561869&	2776786&	1405.04 &   \ovalbox{93320}   &   \ovalbox{802500}  &   316.24    \\
$pipesworld17$&  156733&	561923&	292.99&	\ovalbox{89829}&	334833&	188.93&	93169&	\ovalbox{324196}&	140.06	\\
$pipesworld18$&  21164&	230257&	1049.83&	491&	5057&	30.25&	 491&	\ovalbox{2948}&	28.16 \\
$pipesworld20$&	-	&	-	&	-	&	-	&-	&-	&	\ovalbox{1417}&	\ovalbox{30469}&	596.39 \\
$pipesworld21$&	4486&	29504&	32.43& -	&-	&- & 	\ovalbox{896}&	\ovalbox{4412}&	8.38\\
% DONE
\hline

% DONE
$tpp8     $   &   675&	4987&	0.12&	\ovalbox{570}&	\ovalbox{3854}&	0.14&	581&	4735&	0.16   \\
$tpp9     $   &   \ovalbox{570}&	\ovalbox{3839}&	0.09&	960&	7430&	0.28&	1184&	10098&	0.29   \\
$tpp10    $   &   585&	5861&	0.13&	3679&	37791&	1.18&	\ovalbox{442}&	\ovalbox{4186}&	0.14    \\
$tpp11    $   &   6294&	68216&	3.21&	3815&	33688&	1.93&	\ovalbox{2678}&	\ovalbox{28276}&	1.75   \\
$tpp12    $   &   12922&	139267&	7.42&	6888&	63805&	3.88&	\ovalbox{1947}&	\ovalbox{22579}&	1.56    \\
$tpp13    $   &   7757&	91630&	6.8&	5554&	70013&	5.49&	\ovalbox{4555}&	\ovalbox{58212}&	4.68   \\
$tpp14    $   &   12291&	1705578&	190.82&	15700&	211138&	22.96&	\ovalbox{6054}&	\ovalbox{76611}&	8.02   \\
$tpp15    $   &   9203&	112319&	15.12&	13788&	181676&	29.22&	\ovalbox{3149}&	\ovalbox{36960}&	4.82  \\
$tpp16    $   &		-	&	-	&	-	&	48549&	783502&	176.84&	\ovalbox{17808}&	\ovalbox{256920}&	61.04\\
$tpp20    $   &		-	&	-	&	-	&	-	&-	&-	& \ovalbox{278126}&	\ovalbox{4503504}&	1309.18 \\
% DONE
\hline

$storage13 $  &   \ovalbox{1353}&	\ovalbox{4606}&	0.18&	1354&	4607&	0.25&	1634&	5645&	0.68    \\
$storage14 $  &   265&	2727&	0.2&	212&	2137&	0.19&	\ovalbox{172}&	\ovalbox{1829}&	0.26    \\
$storage15 $  &   57&	\ovalbox{603}&	0.13&	59&	622&	0.13&	\ovalbox{56}&	629&	0.18   \\
$storage16 $  &   52&	697&	0.25&	31&	412&	0.21&	\ovalbox{26}&	\ovalbox{357}&	0.22  \\
$storage17 $  &   \ovalbox{308}&	\ovalbox{4128}&	0.63&	337&	4480&	0.77&	339&	4553&	1.15   \\
$storage18 $  &   361&	5628&	1.68&	482&	6173&	1.86&	\ovalbox{346}&	\ovalbox{5402}&	2.42  \\
$storage19 $  &	  \ovalbox{133171}&	\ovalbox{1426180}&	594.48	&	133771&	1442383&	607.67&	133425&	1436330&	619.37 \\
% DONE
\hline
$ truck1 $&	22&	230&	0.01	&	22&	230&	0.01&	22&	\ovalbox{74}	&	0.01	\\
$ truck2 $&327&	4434&	0.05&	327&	4434&	0.07&	\ovalbox{113}&	\ovalbox{298}&	0.03	\\
$ truck3 $&35&	672&	0.05&	35&	672&	0.05&	\ovalbox{34}&	\ovalbox{148}&	0.05	\\
$ truck4 $&40&	931&	0.02&	40&	931&	0.03&	40&	\ovalbox{227}&	0.02	\\
$ truck5 $&37&	1255&	0.05&	37&	1255&	0.05&	\ovalbox{35}&	\ovalbox{244}&	0.05	\\
$ truck6 $& 47&	1822&	0.08&	47&	1822&	0.09&	\ovalbox{41}&	\ovalbox{255}&	0.07	\\
$ truck7 $&23913&	525066&	9.24&	23913&	525066&	9.76&	23914&	\ovalbox{80585}&	8.32	\\
$ truck8 $&300&	9217&	0.24&	300&	9217&	0.23&	\ovalbox{257}&	\ovalbox{798}&	0.17	\\
$ truck9 $  &   3390&	154042&	2.87&	3390&	154042&	2.95&	3390&	154042&	2.95 \\
$ truck10 $ &   139416&	8111005&	179.09&	139416&	8111005&	186.29&	\ovalbox{86993}&	\ovalbox{218926}&	46.46 \\
$ truck11 $ &   4715&	344079&	13.89&	4715&	344079&	14.25&	\ovalbox{4526}&	\ovalbox{12044}&	8.38 \\
$ truck12 $ &   320765&	30119993&	1357.29&	320765&	30119993&	1352.82&	320790&	\ovalbox{1068035}&	518.89 \\
$ truck13 $ &   525967&	27598625&	1181.73&	525967&	27598625&	1182.49&	\ovalbox{289825}&	\ovalbox{698071}&	349.43 \\
$ truck14 $ &   37380&	863543&	46.23&	37380&	863543&	47.36&	\ovalbox{20713}&	\ovalbox{496998}&	36.68\\
$ truck15 $ &   59659&	1356218&	79.9&	59659&	1356218&	84.26&	\ovalbox{57584}&	\ovalbox{148886}&	62.78 \\
$ truck17 $ &  	-	&	-	&	-	&	-	&-	&-	& 	\ovalbox{159519}&	\ovalbox{452932}&	553.03 \\
% DONE
\hline
$ rover7 $ &  30&	464&	0.01&	30&	464&	0.01&	\ovalbox{20}&	\ovalbox{224}&	0.01 \\
$ rover8 $ & 36&	866&	0.01&	36&	866&	0.02&	\ovalbox{28}&	\ovalbox{554}&	0.01\\
$ rover9 $ & 281&	5865&	0.06&	368&	8349&	0.11&	\ovalbox{113}&	\ovalbox{1148}&	0.02 \\
$ rover10 $ & 232&	7321&	0.09&	232&	7321&	0.12&	\ovalbox{81}&	\ovalbox{1186}&	0.03 \\
$ rover11$ & 1074&	24425&	0.28&	1074&	24425&	0.35&	\ovalbox{418}&	\ovalbox{7754}&	0.13 \\
$ rover12 $ & 26&	531&	0.01&	26&	531&	0.01&	26&	\ovalbox{438}&	0.01 \\
$ rover13 $ & 175&	4523&	0.09&	175&	4523&	0.12&	191&	\ovalbox{3685}&	0.1 \\
$ rover14 $ & 81&	2075&	0.03&	81&	2075&	0.04&	215&	2424&	0.06 \\
$ rover15 $ & 570&	16587&	0.25&	570&	16587&	0.29&	\ovalbox{499}&	\ovalbox{15850}&	0.17 \\
$ rover16 $ & 545&	13945&	0.24&	545&	13945&	0.27&	\ovalbox{103}&	\ovalbox{1771}&	0.05 \\
$ rover17 $ & -  &  -    &  -   &  	445&	16811&	0.54&	\ovalbox{172}&	\ovalbox{2319}&	0.11 \\
$ rover18 $ & 37590&	1532969&	52.87&	37590&	1532969&	63.4&	\ovalbox{4801}&	\ovalbox{110878}&	4.77 \\
% DONE
\hline

$openstack5$  &  39&	171&	0.01 & 39	& 171&	0.01&	39&	\ovalbox{170}&	0.01   \\
$openstack6$  &  109&	712&	0.05&	109&	712&	0.06&	109&	712&	0.08\\
$openstack7$  &   100&	673&	0.05&	100&	673&	0.06&	100&	\ovalbox{672}&	0.09  \\
% DONE
\hline
$pathway3$  &28&	344&	0.01 &	28&	344&	0.01&	\ovalbox{22}&	\ovalbox{84}&	0.01\\
$pathway4$  &76&	726&	0.01&	76&	726&	0.03&	\ovalbox{61}&	\ovalbox{624}&	0.01\\
$pathway5$  &47&	1296&	0.02&	47&	1296&	0.03&	\ovalbox{37}&	\ovalbox{1061}&	0.02	\\
$pathway6$  &255&	6402&	0.12&	255&	6402&	0.28&	\ovalbox{68}&	\ovalbox{626}&	0.04	\\
$pathway7$  &21147&	750778&	20.76&	21147&	750778&	47.21&	21147&	\ovalbox{533384}&	17.28	\\
$pathway8$  &\ovalbox{21507}&	858174&	24.93&  -  &  -    &  -   &	39088&	\ovalbox{472722}&	31.25	\\
$pathway9$  &-  &  -    &  -   &	471&	13639&	1.39&	\ovalbox{151}&	\ovalbox{2155}&	0.19	\\
\hline \hline
\end{longtable}

\end{landscape}

\nop{
\begin{landscape}
\clearpage
\setlength{\LTcapwidth}{\textwidth}
\setlongtables
\input{table3}
\end{landscape}
}

\end{document}